\documentclass[lettersize,journal]{IEEEtran}
\usepackage{amsmath,amsfonts}
\usepackage{array}
\usepackage[caption=false,font=normalsize,labelfont=sf,textfont=sf]{subfig}
\usepackage{textcomp}
\usepackage{stfloats}
\usepackage{url}
\usepackage{verbatim}
\usepackage{graphicx}
\usepackage{cite}
\usepackage{bm}  
\usepackage[ruled,vlined]{algorithm2e}
\usepackage{booktabs} 
\usepackage{amsmath}
\usepackage{multirow}
\PassOptionsToPackage{table,dvipsnames}{xcolor}
\usepackage[table]{xcolor}
\usepackage{colortbl}
\usepackage[accsupp]{axessibility}  
\definecolor{rou}{HTML}{b38270}
\definecolor{mygray}{gray}{.9}
\usepackage{hyperref}
\usepackage{caption}
\usepackage{makecell}
\newcommand{\todo}[1]{\textcolor{red}{\textbf{\tiny{todo}}}}
\newcommand{\neww}[1]{\textcolor[rgb]{0.0, 0.0, 0.0}{#1}}
\newcommand{\newm}[1]{\textcolor[rgb]{0.0, 0.0, 0.0}{#1}}
\newcommand{\newx}[1]{\textcolor[rgb]{0.0, 0.0, 0.0}{#1}}

\begin{document}

\title{DrawMotion\hspace{0.2em}\raisebox{-1mm}{\includegraphics[scale=0.12]{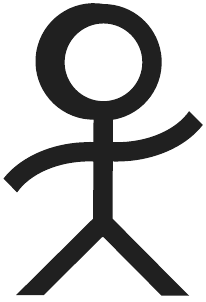}}: Generating 3D Human Motions by Freehand Drawing}

\author{Tao Wang$^{1}$,  Lei Jin$^{\dag1}$\thanks{\textsuperscript{\dag}Lei Jin is corresponding author.}, Zhihua Wu$^{2}$, Qiaozhi He$^{3}$, Jiaming Chu$^{1}$, \\ Yu Cheng$^{4}$, Junliang Xing$^{5}$, Jian Zhao$^{6, 7}$, Shuicheng Yan$^{4}$, \emph{Fellow}, \emph{IEEE}, Li Wang$^{1}$\\
{\small $^{1}$Beijing University of Posts and Telecommunications, $^{2}$University of Science and Technology of China,} \\
{\small $^{3}$NLP Lab, School of Computer Science and Engineering, Northeastern University, Shenyang, China, $^{4}$National University of Singapore,} \\
{\small  $^{5}$Tsinghua University, $^{6}$The Institute of AI (TeleAI), China Telecom, $^{7}$Northwestern Polytechnical University} \\
{\tt\small wangtao@bupt.edu.cn, jinlei@bupt.edu.cn, wuzhh01@mail.ustc.edu.cn, qiaozhihe2022@outlook.com,} \\ 
{\tt\small  chujiaming886@bupt.edu.cn,  e0321276@u.nus.edu, jlxing@tsinghua.edu.cn,} \\
{\tt\small jian\_zhao@nwpu.edu.cn, shuicheng.yan@gmail.com, liwang@bupt.edu.cn}
}



\markboth{Journal of \LaTeX\ Class Files,~Vol.~14, No.~8, August~2021}%
{Shell \MakeLowercase{\textit{et al.}}: A Sample Article Using IEEEtran.cls for IEEE Journals}


\maketitle


\begin{abstract}
Text-to-motion generation, which translates textual descriptions into human motions, faces the challenge that users often struggle to precisely convey their intended motions through text alone. To address this issue, this paper introduces \textbf{DrawMotion}, an efficient diffusion-based framework designed for multi-condition scenarios. DrawMotion generates motions based on both a conventional text condition and a novel hand-drawing condition, which provide semantic and spatial control over the generated motions, respectively. Specifically, we tackle the fine-grained motion generation task from three perspectives: 1) \textbf{Freehand drawing condition.} To accurately capture users' intended motions without requiring tedious textual input, we develop an algorithm to automatically generate hand-drawn stickman sketches across different dataset formats. In addition, a 2D trajectory condition is incorporated into DrawMotion to achieve improved global spatial control. 2) \textbf{Multi-Condition Fusion.} We propose a Multi-Condition Module (MCM) that is integrated into the diffusion process, enabling the model to exploit all possible condition combinations while reducing computational complexity compared to conventional approaches. 3) \textbf{Training-free guidance.} Notably, the MCM in DrawMotion ensures that its intermediate features lie in a continuous space, allowing classifier guidance gradients to update the features and thereby aligning the generated motions with user intentions while preserving fidelity. Quantitative experiments and user studies demonstrate that the freehand drawing approach reduces user time by approximately 46.7\% when generating motions aligned with their imagination. The code, demos, and relevant data are publicly available at \url{https://github.com/InvertedForest/DrawMotion}.
\end{abstract}


\section{Introduction}
The task of human motion generation~\cite{zhang2023remodiffuse, zhang2024motiongpt, tevet2022motionclip} has a wide range of applications across diverse fields, including film and television production, virtual reality, the gaming industry, and beyond. Specifically, the popular sub-task of motion generation, text-to-motion, can generate natural human motion sequences based on language descriptions, freeing 3D animators from manually key-framing 3D character poses.

However,  it is evident that a simple description such as ``A high kick forward'' may not fully capture users' detailed imagination of the complex arm gesture shown in Figure~\ref{fig:main}. 
Previous works~\cite{kim2023flame, zhang2024motiongpt, zhang2024finemogen, zhang2022motiondiffuse} focus on generating the desired motion with complex textual descriptions. For instance, Flame~\cite{kim2023flame} allows for appending additional textual descriptions to modify the character's motion sequence based on a diffusion model. FineMoGen~\cite{zhang2024finemogen} controls the individual body parts of the 3D character through detailed descriptions. Goel et al. 2024~\cite{goel2024iterative} propose an intermediate representation (IR) for text-driven kinematic motion edits, which control joint location and rotation with Python code generated from the large language model. 
These approaches improve alignment between generated motions and user intentions by enhancing textual descriptions.  However, user demand for more accurate outputs necessitates more detailed textual descriptions. Based on the above, we propose a novel hand-drawing condition to control the details of human motion sequences and mitigate the need for extensive descriptions.

The proposed hand-drawing condition includes a hand-drawn trajectory and stickman figures specified in the trajectory. This condition greatly reduces the difficulty of precisely generating the motion that the user wants and enhances the user experience during hand-drawing as shown in Figure~\ref{fig:compare}. Unlike our previous work StickMotion~\cite{wang2025stickmotion}, which can only specify 3 frames and dynamically place their positions, DrawMotion allows multiple stickman figures to be inserted at arbitrary positions along the input trajectory. Removing this restriction provides greater flexibility and precision, while also requiring users to be more responsible for the fidelity of the final results.

\begin{figure}[t]
\begin{center}
\includegraphics[width=1.0\linewidth]{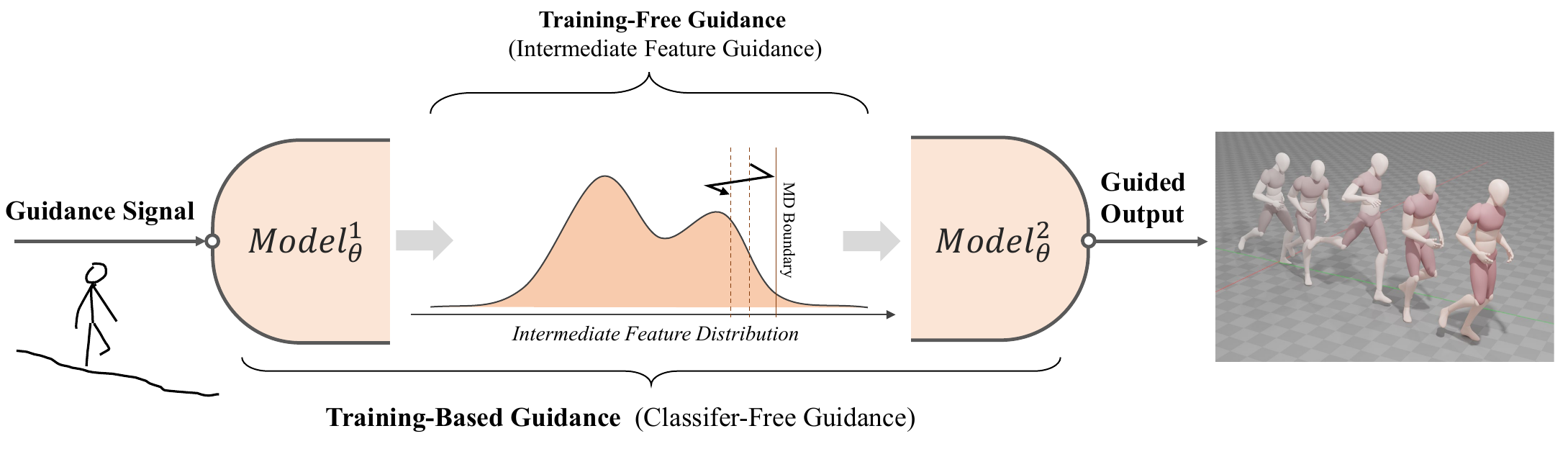}
\end{center}
\caption{\small{Pipeline of DrawMotion inference. In addition to the training-based guidance, a training-free guidance updates the intermediate feature of the model within the MD boundary to ensure that the generations meet the conditions while maintaining its fidelity.}}
\vspace{-2mm}
\label{fig:head}
\end{figure}

Nevertheless, these desired functionalities pose three challenges for DrawMotion: 
1) \emph{Data generation.} Hand-drawn stickman figures are limited by the drawing style of the annotators and are time-consuming to collect. We propose a Stickman Generation Algorithm (SGA) that automatically produces stickman sketches in diverse styles, as shown in Figure~\ref{fig:stickman}.
2) \emph{Multi-Condition Fusion.} Previous works~\cite{chen2022re, zhang2023remodiffuse} achieve all possible combinations of two conditions via the mask operation for condition input in self-attention~\cite{vaswani2017attention, zhang2023remodiffuse} module, but this introduces redundant computation when calculating the masked-token attention. We instead design an efficient Multi-Condition Module (MCM) to process multiple conditions, as detailed in Section~\ref{sec:fusion}.  
3) \emph{Trajectory alignment.} DrawMotion must balance fidelity, text conditions, stickman conditions, and trajectory conditions during the generation. Although trajectory provides the global motion path, text influences global semantics, often counteracting trajectory constraints. To address this, we propose a training-free guidance strategy (Intermediate Feature Guidance, IFG) that improves trajectory alignment by leveraging the continuity of the MCM's intermediate feature space (Figure~\ref{fig:head}).

The main contributions of this work are summarized as follows:
\begin{enumerate}[]  
\item[\textbullet\ ] To the best of our knowledge, we are the first to introduce hand-drawn representations as a condition for motion generation, enabling users to precisely control motion details through simple sketches without extensive textual descriptions.
\item[\textbullet\ ] We propose a Multi-Condition Module (MCM) for condition fusion in the diffusion process, reducing computational complexity while improving performance compared to the standard self-attention module. Different variants of self-attention are applied based on global or local attributes of each condition to enhance consistency between generated results and conditions.
\item[\textbullet\ ] We show that the intermediate feature space of MCM is relatively continuous, enabling us to design a novel training-free guidance method (IFG) that significantly reduces computational overhead while improving fidelity and alignment.
\item[\textbullet\ ] We evaluate DrawMotion on both the KIT-ML and HumanML3D datasets, demonstrating competitive performance with state-of-the-art text-to-motion methods, while achieving superior results in \emph{StiSim} (stickman similarity) and \emph{Traj.Err} (trajectory alignment).
\end{enumerate}


\neww{A preliminary version of this work appeared as StickMotion~\cite{wang2025stickmotion}, which was designed with a primary focus on usability. StickMotion introduced a self-supervised stickman encoding method via SGA and a primary Multi-Condition Module (MCM) to fuse text and stickman conditions, \newx{where} stickman poses are placed at fixed and automatically determined temporal locations to ensure global coherence. While effective, this design inherently provides only \emph{coarse-grained control}, as users cannot precisely specify the spatial trajectory of motion nor arbitrarily constrain poses on the motion sequence.
DrawMotion is motivated by the need for a more \emph{fine-grained and professional control interface}. Compared to StickMotion, this work goes beyond an incremental extension and addresses several fundamental challenges introduced by explicit trajectory control and flexible pose placement. Specifically, we make the following key advances:  
1) DrawMotion incorporates explicit 2D trajectory conditions and allows users to place multiple stickman poses at arbitrary positions along the trajectory. This greatly  increases user control but also requires handling conflicts between text semantics, spatial trajectories, and pose constraints. To address this, we introduce both training-based conditioning and a novel training-free guidance mechanism.  
2) We redesign and refine the MCM by adopting modality-specific condition decoders, enabling more effective fusion of heterogeneous inputs. More importantly, we show that the intermediate features produced by MCM form a continuous and guidance-receptive space, which directly motivates our Intermediate Feature Guidance (IFG). IFG allows strict trajectory alignment at inference time without retraining and with lower computational cost than existing motion editing methods.  
3) We further enhance the stickman representation of the stickman encoder with a candidate loss that preserves multiple plausible pose hypotheses, and we provide extensive quantitative and qualitative evaluations demonstrating that DrawMotion consistently outperforms StickMotion and other state-of-the-art methods in fine-grained, user-controlled motion generation.
Together, these contributions establish DrawMotion not only as a substantial advancement over StickMotion, but also as a strong baseline for interactive and precise human motion generation.
}

The remainder of this paper is organized as follows: Section~\ref{sec:rw} reviews related work; Sections~\ref{sec:tb} and~\ref{sec:tf} introduce our training-based and training-free guidance strategies; Section~\ref{sec:exp} reports experimental results and analyses, and the last two sections conclude the paper.

\section{Related Work}~\label{sec:rw}

\noindent\textbf{Diffusion Models.}
In recent years, significant progress has been made in applying deep learning-based generative models, particularly in diffusion models. The proposed denoising diffusion probabilistic model (DDPM)~\cite{sohl2015deep,ho2020denoising} aims to learn the process of restoring original data that has been corrupted by noise, progressively eliminating the noise during inference and resulting in final outputs that closely approximate the distribution of the original data. ADM~\cite{dhariwal2021diffusion} first achieves superior sample quality compared to Generative Adversarial Networks (GAN)~\cite{goodfellow2020generative} with its proposed Denoising Diffusion Implicit Model (DDIM). ADM also incorporates classifier guidance inspired by GANs to control the categories of generated content. Jonathan Ho and Tim Salimans~\cite{ho2021classifier} propose a classifier-free guidance technique for reducing sample diversity in diffusion models without relying on a classifier. Currently, diffusion models~\cite{cao2024survey} are employed for generating various data types such as images, videos, text, sound, time series data, \emph{etc}.

\noindent\textbf{Human Motion Generation.}
Human motion generation aims to generate natural sequences of human motion based on various forms of control conditions. This task can be categorized into the following types depending on the conditions. 
Motion prediction task~\cite{guo2023back, chen2023humanmac, zhang2024incorporating, ma2022progressively, wang2024gcnext} involves using previous human motion sequences as input to predict the subsequent sequences. This task can be applied to autonomous driving and social security analysis. 
Action-to-motion task~\cite{guo2020action2motion, yu2020structure, degardin2022generative, petrovich2021action, lu2022action, cervantes2022implicit} generates human motion sequences based on specified action categories, providing a more direct but coarse-grained control over human motion. 
Sound-to-motion task can be further divided into music-to-dance~\cite{gao2023dancemeld, huang2020dance, li2022danceformer, tseng2023edge} and speech-to-gesture~\cite{ao2023gesturediffuclip, ghorbani2023zeroeggs, kucherenko2019analyzing, yoon2020speech} tasks, which simultaneously generate corresponding human motions or gestures in response to audio stimuli. 
Text-to-motion task~\cite{ahuja2019language2pose, ghosh2021synthesis, tevet2022motionclip, guo2022generating, cui2024anyskill, zhang2023remodiffuse, zhang2024motiongpt, guo2024momask} generates human motion sequences from natural language descriptions like ``walk fast and turn right'' or ``squat down then jump up''. However, users often struggle to precisely control the position of each limb with limited textual description alone. 
Additionally, there are interaction-to-motion tasks that consider interactions between humans and scenes~\cite{huang2023diffusion, hassan2021populating, lim2023mammos, liu2023revisit, xiao2023unified} / objects~\cite{diller2024cg, xu2023interdiff, gao2020interactgan, lin2023handdiffuse} / humans~\cite{cai2024digital, chopin2024bipartite, ghosh2023remos, liang2024intergen, tanaka2023role}, while incorporating generated human motions as reactions in digital environments.

\noindent\neww{\textbf{Diffusion-based Motion Editing Methods.}}
\neww{
Motion editing with diffusion models has attracted increasing attention, aiming to modify generated motions under user-specified spatial constraints while preserving naturalness. Existing approaches can be broadly categorized into two paradigms:
}

\noindent\neww{\textbf{1) Training-based methods} incorporate spatial constraints during model training or via auxiliary modules. For example, GMD~\cite{Karunratanakul2023GuidedMD} trains separate models for trajectory generation and trajectory-conditioned motion synthesis, and employs classifier guidance to align motions with target trajectories. PriorMDM~\cite{Shafir2023HumanMD} introduces partial-noise training to preserve invariant motion dimensions, providing the model with reliable partial data for motion editing. CondMDI~\cite{Cohan2024FlexibleMI} extends this idea by converting relative root orientations to global coordinates and applying classifier-free guidance, thereby improving trajectory control and motion fidelity. OmniControl~\cite{Xie2023OmniControlCA} combines a base diffusion model with ControlNet~\cite{Zhang2023AddingCC}, integrating auxiliary networks to guide motion generation under spatial and textual conditions, thereby achieving a balanced trade-off between user constraints and motion naturalness. While these methods generally achieve lower FID and Traj.Err., they require additional training or architectural modifications.}

\noindent\neww{\textbf{2) Training-free methods} enforce constraints during inference without modifying model parameters. Diffusion inpainting approaches, such as MDM~\cite{tevet2022humanmotiondiffusionmodel}, directly overwrite noised motion data $x_{t-1}$ at specified positions during each denoising step. However, this strategy disrupts the natural distribution of $x_{t-1}$, and the model may interpret the injected values as noise and discard them. Classifier guidance methods, adopted in GMD~\cite{Karunratanakul2023GuidedMD}, OmniControl~\cite{Xie2023OmniControlCA}, and DNO~\cite{Karunratanakul2023OptimizingDN}, backpropagate spatial losses to $x_{t-1}$, $x_T$, or intermediate features to steer the generation process. Although these methods improve alignment with user constraints, they may reduce motion vividness and often struggle with sparse or conflicting spatial supervision. DNO further optimizes the initial noise through multiple gradient backpropagations, achieving constraint satisfaction at the cost of significantly increased computational overhead.}

\neww{In practice, combining training-based and training-free strategies, as in OmniControl~\cite{Xie2023OmniControlCA} and DrawMotion, often yields a better balance between constraint alignment and motion naturalness. Compared to purely training-free methods, these hybrid approaches achieve superior Traj.Err. and FID, demonstrating the effectiveness of integrating training-based and training-free paradigms.}

\section{Training-Based Guidance}\label{sec:tb}

\noindent\textbf{Overview.} 
DrawMotion leverages both hand-drawn sketches and textual descriptions as input modalities. Users may provide any combination of these two modalities, \emph{i.e.}, $C(\text{text}, \text{draw})$, $C(\text{text}, \varnothing)$, $C(\varnothing, \text{draw})$, and $C(\varnothing, \varnothing)$. This section is structured as follows: Section~\ref{sec:stickre} introduces our method for generating hand-drawing representations without manual annotation; Section~\ref{sec:diffusion} provides a concise overview of the general classifier-free guidance framework based on diffusion models, which we adopt in our approach; Section~\ref{sec:fusion} then presents our proposed Multi-Condition Module (MCM), which improves upon traditional multi-condition fusion techniques and naturally leads into Section~\ref{sec:tf} for the proposed training-free guidance.

\subsection{Hand-Drawing Representation}\label{sec:stickre}
User-provided hand-drawn sketches consist of trajectories and stickman figures. We stipulate that a hand-drawing representation must include one trajectory, while any number of stickman figures can be placed along it. It is therefore crucial to address the challenges of generating, encoding, and applying such representations.

\noindent\textbf{\neww{2D} Trajectory.}
\neww{After the user draws a 2D trajectory on the web interface, the frontend returns a coordinate sequence $J^{t} \in \mathbb{R}^{(n, 2)}$, where $n$ denotes the number of sampled points. The trajectory is then resampled to $\widehat{J}^{t} \in \mathbb{R}^{(T, 2)}$, where $T$ represents the target number of motion frames. The resampling process can be biased toward uniform resampling (ignoring drawing speed) or density-based resampling (preserving drawing speed), and the trajectory can be freely transformed according to the user’s intent. The resampled trajectory $\widehat{J}^{t}$ is subsequently fed into DrawMotion as the target 2D pelvis path, enabling fine-grained control over both motion trajectory and speed.
}

The above details the trajectory processing at inference time. During training, trajectories from motion sequences in the dataset are directly input into DrawMotion, with additional supervision applied as shown in Equation~\ref{eq:samp}. The reason for directly using hand-drawn trajectories as input is twofold:  
1) Both hand-drawn and real motion trajectories exhibit inertia: the former reflects the inertia of the hand, while the latter reflects the inertia of the human body. After density-based sampling, the two align in terms of inertial characteristics.  
2) As illustrated in Figures~\ref{fig:vis} and~\ref{fig:compare}, DrawMotion fine-tunes the trajectory of the generated motion sequence to ensure high fidelity and consistency with the trajectory condition. This enables the model to incorporate the imperfect hand-drawn trajectories as effective guidance.

\begin{figure}[t]
\begin{center}
\includegraphics[width=0.9\linewidth]{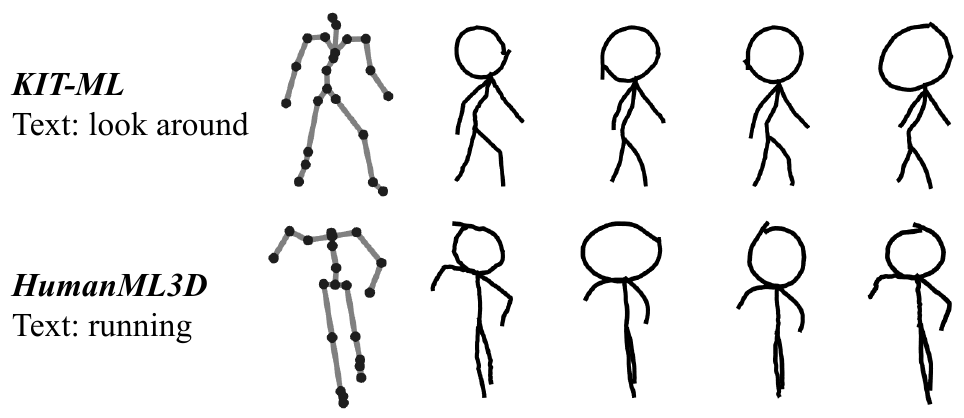}
\end{center}
\vspace{-4mm}
\caption{\small{Stickmen generated by Stickman Generation Algorithm on the KIT-ML~\cite{plappert2016kit} and HumanML3D~\cite{guo2022generating} datasets.}}
\vspace{-2mm}
\label{fig:stickman}
\end{figure}

\noindent\textbf{Stickman Generation Algorithm.}  Due to the lack of hand-drawn stickmen in existing datasets, we propose a Stickman Generation Algorithm (SGA) based on the 3D coordinates of human joints from existing motion datasets to automatically generate hand-drawn stickmen. Considering the characteristics of human hand-drawing, we take into account the following aspects: 
   1) \emph{Stroke smoothness.} The smoothness of strokes is influenced by force and individual preferences. Moreover, the smoothness of drawing trajectories may vary across different devices. For instance, strokes drawn with a mouse tend to be more jittery than those created on an iPad.
   2) \emph{Misplacement.} Inevitably, inaccuracies in pen placement may lead to global positional deviations in these body parts.
   3) \emph{Scaling.} Hand-drawings focus on local details while disregarding global information, resulting in size discrepancies among different body parts.
The stickmen generated from different datasets  are shown in Figure~\ref{fig:stickman}. Moreover, the stickman may appear similar when observing different poses from various angles, so we stipulate that the stickman should be obtained by observing the human pose from the front, \emph{i.e.}, where the line of sight is approximately perpendicular to the pelvic plane of the pose.

\noindent\textbf{Information Encoding.}  
A trade-off exists between user convenience and computational efficiency when processing stickman information. A direct approach would require collecting at least 200 two-dimensional coordinate points (estimated from visualization) with connectivity information to faithfully reconstruct the drawing. However, this incurs high memory and computational cost due to the $200^2$ pairwise interactions among points. To reduce overhead, we propose a compact representation in which users draw six one-stroke lines representing the head, torso, and four limbs in any order. Each line is individually encoded and then aggregated by a transformer encoder~\cite{vaswani2017attention} to produce a stickman embedding. This compact encoding reduces computational complexity while improving recognition accuracy.  

\noindent\textbf{Stickman Encoder.}  
Pre-training and freezing the stickman encoder significantly enhances DrawMotion's performance. To this end, we train an autoencoder consisting of a stickman encoder and a feature-to-pose decoder. The encoder maps stickmen into embeddings, while the decoder reconstructs the original pose from these embeddings, preserving pose information. The decoder predicts $N$ candidate 3D poses with the following loss:  
\begin{equation} 
\begin{aligned}
& \ell_n = 0.1 \times \lVert \text{limb\_offset}^\text{gt} - \text{limb\_offset}^\text{pred}_n \rVert_2^2, \\
& \ell^{\text{final}} = 10 \times \ell_k + \sum_{n=1}^{N} \ell_n,\ \text{where}\ k = \underset{n}{\arg\min} \, \ell_n,
\end{aligned}\label{eq:cand}
\end{equation}
where $\text{limb\_offset}$ denotes the 3D offset between adjacent joints. The candidate loss is motivated by two factors:  
1) When two limbs (e.g., arms or legs) are close together, stickmen often cannot be reliably distinguished between left and right (see the second row of Figure~\ref{fig:stickman}).  
2) Pose estimation from stickmen, whether from algorithmic generation or user sketches, inevitably introduces noise.  
Thus, forcing the decoder to predict a single exact pose may result in latent information loss and ambiguous outputs. The candidate loss alleviates this problem and improves motion prediction accuracy, as demonstrated in Table~\ref{tab:t2m}.  

\noindent\neww{
\textbf{Trajectory Encoder.} 
Unlike \newx{the} stickman encoder, we do not pretrain the trajectory encoder. Instead, it is trained jointly with the entire DrawMotion model. This is because the trajectory information is relatively direct, with each point representing the pelvis position. 
Specifically, the trajectory encoder consists of six Conv1d layers with activation functions. The trajectory $\widehat{J}^{t} \in \mathbb{R}^{(T,2)}$ is encoded to the trajectory encoding $e^j \in \mathbb{R}^{(T,E)}$, here $T$ denotes the motion sequence length, and $E$ represents the channel dimension of the encoding.
}

\begin{figure*}[t]
\begin{center}
\includegraphics[width=1.0\linewidth]{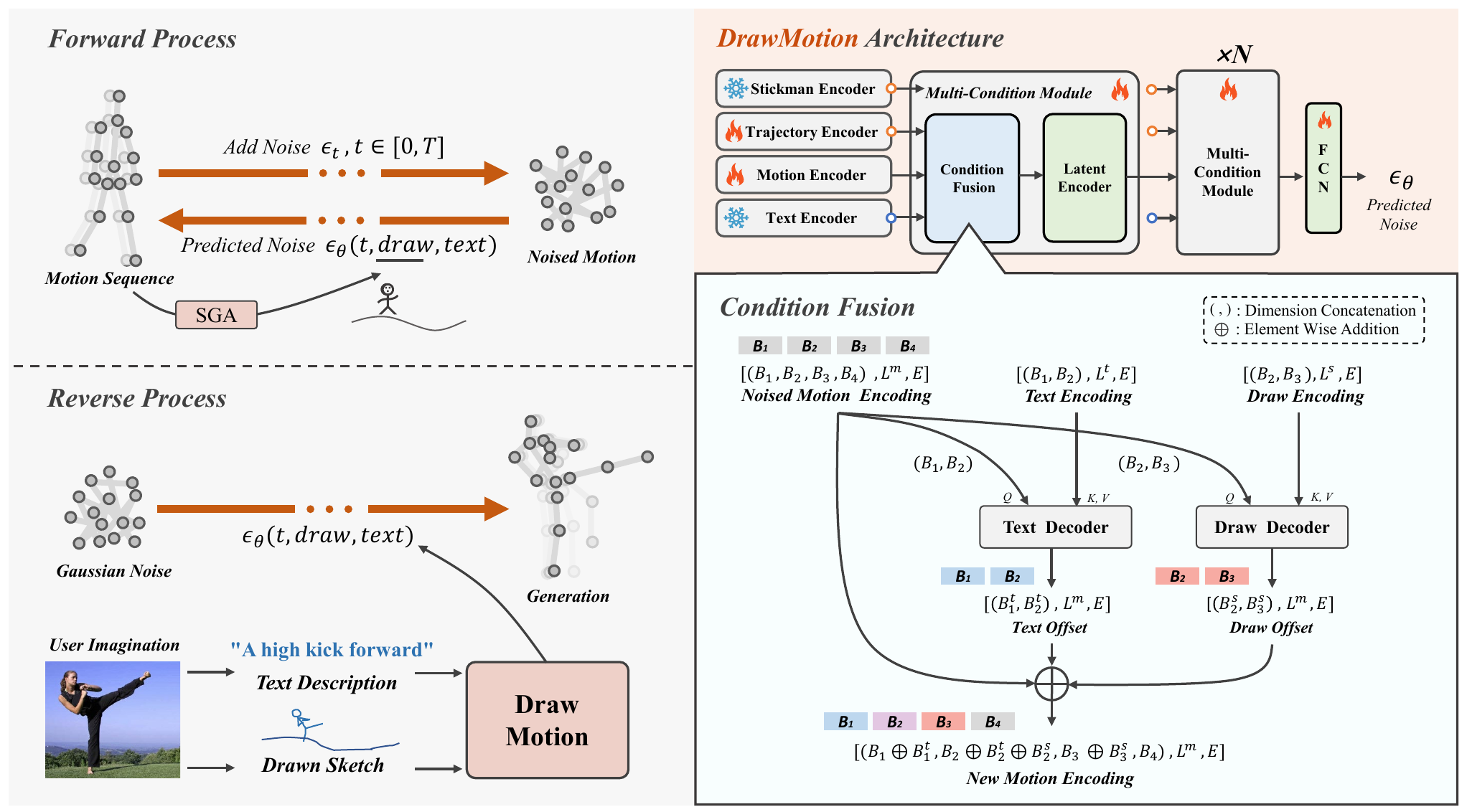}
\end{center}
\vspace{-2mm}
\caption{\small{The DrawMotion framework consists of the diffusion process (left) and the network structure (right). 
1) The diffusion process includes a forward and a reverse process. In the forward process, original motions are augmented with Gaussian noise and fed into DrawMotion, which learns to predict the added noise based on textual descriptions and hand-drawn sketches. In the reverse process, user-provided textual descriptions and hand-drawn sketches are input into DrawMotion, enabling the gradual generation of motion sequences using the predicted noise.  
2) In the DrawMotion architecture, both the stickman encoder and the text encoder are frozen, while the remaining modules are trainable. Encoded inputs are processed by multiple Multi-Condition Modules (MCMs) to get the final output.}}
\label{fig:main}
\end{figure*}

\subsection{Diffusion-based Motion Generation}\label{sec:diffusion}
Diffusion-based works have demonstrated excellent performance in the field of human motion generation. We adopt diffusion models as the base model for DrawMotion because diffusion models can control the bias towards generating motions based on either textual description or hand-drawing conditions.  
Diffusion models aim to approximate the data distribution $q(x_0)$ with a model distribution $q_\theta(x_0)$, where $\theta$ denotes the learnable parameters of DrawMotion.
As illustrated in Figure~\ref{fig:main}, the process consists of two stages: a forward (noising) process and a reverse (denoising) process.  

\noindent\textbf{Forward process.}  
In the forward process, Gaussian noise $\epsilon_t \sim \mathcal{N}(0, \mathbf{I})$ is gradually added to the clean motion $x_0$ over timesteps $t \in [0, T]$, using a variance schedule $\beta_t$. The process is defined as  
\begin{equation} 
\begin{aligned}
&q(\mathbf{x}_{1:T}|\mathbf{x}_0) = \prod_{t=1}^T q(\mathbf{x}_t|\mathbf{x}_{t-1}),\\
&q(\mathbf{x}_t|\mathbf{x}_{t-1}) = \mathcal{N}\!\left(\mathbf{x}_t;\sqrt{\alpha_t}\mathbf{x}_{t-1}, (1-\alpha_t)\mathbf{I}\right),
\end{aligned}\label{eq:forward}
\end{equation}
which is equivalent to  
\[
\mathbf{x}_t = \sqrt{\bar{\alpha}_{t}}\,\mathbf{x}_0 + \sqrt{1-\bar{\alpha}_{t}}\,\epsilon_t,\quad \bar{\alpha}_t=\prod_{s=1}^t\alpha_s.
\]  
Thus, $x_t$ can be sampled directly from $x_0$ without iteratively generating intermediate states. When $t=T$, $x_T \sim \mathcal{N}(0, \mathbf{I})$.  

During training, DrawMotion minimizes the following denoising objective:  
\begin{equation}
\mathbb{E}_{\epsilon_t,t,x_0}\!\left[\left\lVert \epsilon_t-\epsilon_\theta(\mathbf{x}_t,t,L,C(\text{draw}), C(\text{text}))\right\rVert^2\right],
\label{eq:motion_loss}
\end{equation}
where $L$ is the sequence length, $C(\text{draw})$ and $C(\text{text})$ are the drawing and text conditions (activated with probabilities $p^c_{\text{draw}}$ and $p^c_{\text{text}}$, both set to 0.7), and $\epsilon_\theta(\cdot)$ denotes the noise predictor.  

\noindent\textbf{Reverse process.}  
In the reverse process, starting from $x_T \sim \mathcal{N}(0, \mathbf{I})$, the model gradually removes noise to recover realistic motion sequences. According to DDPM~\cite{ho2020denoising}, the reverse transition is parameterized as  
\begin{equation} 
\begin{aligned}
p_\theta(x_{t-1}|x_t) &= \mathcal{N}\!\bigl(\mu_t(\epsilon_{\theta}, x_t),\ \sigma_t^2 \mathbf{I}\bigr),\\
\mu_t(\epsilon_\theta, x_t) &= \tfrac{1}{\sqrt{\alpha_t}} \Bigl( x_t - \tfrac{1 - \alpha_t}{\sqrt{1 - \bar{\alpha}_t}} \,\epsilon_\theta(x_t, t) \Bigr),
\end{aligned}\label{eq:reverse}
\end{equation}
where $\epsilon_\theta$ is the predicted noise and $\sigma_t$ is the variance coefficient.  

DDIM~\cite{Song2020DenoisingDI} further introduces a deterministic variant that accelerates sampling and improves controllability:  
\begin{equation} 
\begin{aligned}
x_{t-1} &= \hat{\mu}_t(\epsilon_\theta, x_t) + \sqrt{1-\alpha_{t-1}}\, \epsilon_\theta(x_t, t), \\
\hat{\mu}_t(\epsilon_\theta, x_t) &= \sqrt{\alpha_{t-1}} \left( \frac{x_t - \sqrt{1-\alpha_t}\,\epsilon_\theta(x_t, t)}{\sqrt{\alpha_t}} \right),
\end{aligned}\label{eq:ddim}
\end{equation}
where $\hat{\mu}_t$ represents the predicted clean motion $x_0$. We adopt DDIM for the reverse process due to its efficiency and stability. Unlike DDPM, DDIM allows for non-random, deterministic sampling paths, which can reduce the number of steps needed to generate high-quality sequences.

\noindent\textbf{Condition mixture.}  
To bias the denoising process toward different condition combinations, we compute a weighted mixture of predicted noises:  
\begin{equation}
\begin{aligned}
\hat{\epsilon}_{\theta} =\ &w_1 \cdot \epsilon_{\theta}(\text{text}, \text{draw})+w_2 \cdot \epsilon_{\theta}(\varnothing, \text{draw})\\
&+ w_3 \cdot \epsilon_{\theta}(\text{text}, \varnothing)+w_4 \cdot \epsilon_{\theta}(\varnothing, \varnothing).
\end{aligned}\label{eq:mixture}
\end{equation}

Adhering to the principle $w_1+w_2+w_3+w_4=1$~\cite{ho2021classifier} that preserves output statistics, we propose an efficient condition mixture by considering the characteristics of drawing and text conditions inspired by previous works~\cite{chen2022re, zhang2023remodiffuse}. Initially, during the time interval $t \in [T, T/10]$, the approximate motion sequence is determined, and the condition mixture follows the formula $(w_1=w$, $w_2=\hat{w}$, $w_3=w-\hat{w}, w_4=1-2\cdot w)$. Here, 1) constant $w>1$~\cite{ho2021classifier} adjusts the condition sampling strength; 2) $w_1=w$ ensures that the fusion of drawing and text is harmonious. 3) $p(\hat{w}=w) + p(\hat{w}=0) = 1$, with $p(\hat{w}= w)$ controlling the preference of generated motion for the hand-drawing condition. 4) $w_4=1-2\cdot w$ controls the constant distribution of the output. In the final stage ($t \in [T/10, 0]$), we set $(w_1=1, w_{2,3,4}=0)$ to use all conditions to further refine the preliminary result from the beginning stage. Finally, a motion sequence corresponding to the hand-drawing condition and text condition is generated in the reverse process.

\subsection{Architecture of DrawMotion}\label{sec:network}
The diffusion model offers DrawMotion a straightforward training and controllable generation process. However, designing a network architecture that efficiently handles both hand-drawing and text conditions for the diffusion process is equally crucial. As depicted in the right section of Figure~\ref{fig:main}, DrawMotion comprises four input encoders and Multi-Condition Modules (MCMs) to generate the final predicted motion. The input encoders transform the noisy motion, text, and hand-drawing into vectors, which are subsequently fed into  MCMs to produce the final outputs under multiple condition combinations, \emph{i.e.},  {\small$(\text{text}, \text{draw}), (\text{text}, \varnothing), (\varnothing, \text{draw})$}, and {\small$(\varnothing, \varnothing)$}.

\noindent\textbf{Input.}
The input data consists of noisy motion sequences, trajectories, stickman figures, and textual descriptions, which are encoded into $e^m, e^j, e^s, e^t$ with dimensions $[\neww{T},E], [\neww{T},E], [\neww{T},E]$, and $[L,E]$, respectively. Here, \neww{$T$ denotes the motion sequence length, } $L$ denotes the length of the input text encoding, and $E$ represents the dimension of each token in the encoding. Specifically, a simple linear layer is employed to encode the motion sequences; a 1D convolutional neural network (1D CNN) is used to encode the sampled trajectories; CLIP ViT-B/32 \cite{clip}, containing 154 million parameters, is utilized for textual encoding; and a standard transformer encoder (as presented in Section \ref{sec:stickre}) is leveraged for encoding stickman figures.

\noindent\textbf{Condition Decoder Structure.}
The interaction between the input representation and the motion feature is realized through \neww{two kinds} \newx{of} cross-attention mechanisms \neww{based on the different properties of the inputs}. Specifically, the query is derived from the motion sequences with token length $n$, while the key and value are obtained from the condition representations with token length $m$. If we denote the queries, keys, and values as matrices $\boldsymbol{Q} \in \mathbb{R}^{n \times d_k}$, $\boldsymbol{K} \in \mathbb{R}^{m \times d_k}$, and $\boldsymbol{V} \in \mathbb{R}^{m \times d_v}$, respectively, the attention mechanism operates differently depending on the type of condition:

1) Draw Decoder \neww{(standard attention)}. The stickman \neww{$e^s$} and trajectory encoding \neww{$e^j$} \newx{determine} the local human pose and global spatial positioning of each frame in the generated motion sequence $e^m$. The attention mechanism is defined as:
\neww{
\begin{equation} 
\begin{aligned}
&e^{kv} = \text{concat}((e^m \oplus e^j), e^s),\\
&\boldsymbol{Q} = FCN_1(e^m), \boldsymbol{K}, \boldsymbol{V} =  FCN_{2,3}(e^{kv}),\\
&\boldsymbol{D}(\boldsymbol{Q}, \boldsymbol{K}, \boldsymbol{V}) = \text{softmax}\left(\boldsymbol{Q}\boldsymbol{K}\right)\boldsymbol{V},
\end{aligned}\label{eq:att}
\end{equation}
}

\noindent\neww{where ``$\oplus$'' denotes element-wise addition, and ``$\text{concat}$'' denotes concatenation along the sequence (token) dimension. The embeddings $e^{m}$, $e^{j}$, and $e^{s} \in \mathbb{R}^{(T, E)}$ are combined to form $e^{kv} \in \mathbb{R}^{(2 \times T, E)}$.} This attention mechanism, known as dot-product attention~\cite{vaswani2017attention}, is widely used. \neww{It is particularly well suited to the Draw Decoder, as it explicitly models interactions among all tokens. Since the stickman and trajectory conditions encode frame-wise local poses and global spatial coordinates, respectively, the attention mechanism enables motion queries to identify and focus on drawing information corresponding to their respective frames. Meanwhile, cross-frame interactions naturally preserve temporal consistency by allowing each frame to attend to its surrounding context.}


2) Text Decoder \neww{(efficient attention)}. The text representation $e^t$ controls the global semantics of the generated motion sequence $e^m$. Given its global nature, we employ efficient attention~\cite{shen2021efficient}, formulated as:
\begin{equation} 
\begin{aligned}
&\boldsymbol{Q} = \text{softmax}\left(FCN_4(e^m)\right),\\
&\boldsymbol{K}, \boldsymbol{V} =  FCN_{5,6}(\neww{\text{concat}}(e^m, e^t)),\\
&\boldsymbol{D}(\boldsymbol{Q}, \boldsymbol{K}, \boldsymbol{V}) = \boldsymbol{Q} \cdot \left(\text{softmax}\left(\boldsymbol{K}^{\intercal}\right)\boldsymbol{V}\right).
\end{aligned}\label{eq:eff}
\end{equation}
In this formulation, $\boldsymbol{K}$ and $\boldsymbol{Q}$ first learn a \neww{channel} mapping from $d_v$ to $d_k$, capturing global semantic information, which is then mapped sequentially to each query token. \neww{Here, efficient attention not only aligns well with the global semantic nature of textual information, but also significantly reduces computational cost, since its complexity scales linearly with the query token length $n$.} 

The selection of the above two attention mechanisms is our best practice. For ablation experiments, please refer to Section~\ref{sec:abl_decoder}. 
Additionally, we employ efficient attention for the Latent Encoder in MCM as shown in Figure~\ref{fig:main} to further reduce computational complexity.

\noindent\textbf{Multi-Condition Module.}\label{sec:fusion}
Condition combinations are essential for the diffusion process to fuse these conditions. Traditional methods utilize a mask mechanism based on a single self-attention layer to achieve condition combinations. This approach allows the $\boldsymbol{K}$ and $\boldsymbol{V}$ to contain two types of condition information along the token dimension (with $\boldsymbol{Q}$ derived from the input motion). When conducting condition combinations, the attention weights at the positions corresponding to unwanted conditions in the attention map are masked out along the token dimension, preventing the network from perceiving those conditions. The mask mechanism not only wastes computational resources but also restricts the integration of different conditions and inputs, due to its uniform attention structure as mentioned above.

These combinations are implemented efficiently through the Multi-Condition Module (MCM) in DrawMotion.  Within each MCM, a Condition Fusion module is utilized to incorporate hand-drawing and text conditions into the motion feature in the latent space as shown in Figure~\ref{fig:main}. Subsequently, the modified motion feature undergoes re-encoding by the Latent Encoder for further fusion.
Specifically, we partition all data along the batch dimension into four segments ({\small$B_1, B_2, B_3, B_4$}), representing four combinations of text and hand-drawing conditions,  \emph{i.e.},  {\small$(\text{text}, \text{draw}), (\text{text}, \varnothing), (\varnothing, \text{draw})$}, and {\small$(\varnothing, \varnothing)$}. 
In the Condition Fusion module, the Text Decoder and Draw Decoder process only the text input and drawing input for batches ({\small$B_1,B_2$}) and ({\small$B_2,B_3$}), respectively.
By summing up these predicted offsets with their corresponding motion feature along the batch dimension, we obtain new motion features for three condition combinations with only two condition decoders as shown in Figure~\ref{fig:main}. Subsequently, the Latent Encoder re-encodes the fused features for further integration. Compared with conventional approaches~\cite{zhang2023remodiffuse} that rely on masked self-attention, MCM reduces computational complexity and improves DrawMotion's performance (see Section~\ref{sec:abl_mcm}). Moreover, MCM provides the foundation for the training-free guidance method discussed in Section~\ref{sec:tf}.
 
\subsection{Supervision}\label{sec:loss}
To train DrawMotion effectively under multiple condition settings, we design a unified supervision objective.  
As shown in Equation~\ref{eq:samp}, the overall loss integrates three components: trajectory loss, stickman loss, and motion reconstruction loss. Here, $x$ denotes the ground-truth motion sequence and $\hat{x}$ represents the prediction from DrawMotion. The operator $\text{Traj}(\cdot)$ extracts the global trajectory from a motion sequence, while $\text{Pose}(\cdot)$ converts a frame of a motion sequence into a 3D pose.  

\begin{equation} 
\begin{aligned}
 \mathcal{L}_{\text{traj}} &= \big\|\text{Traj}(\hat{x}(\text{draw},*)) - \text{Traj}(x)\big\|_2^2, \\
 \mathcal{L}_{\text{stick}} &= \frac{1}{M} \sum_{i=0}^{L} m_i \cdot \big\|\text{Pose}(\hat{x}_i(\text{draw},*)) - \text{Pose}(x_i)\big\|_2^2, \\
 \mathcal{L}_{\text{motion}} &= \sum_{l=0}^{L} \big\|\hat{x}_l(*,*) - x_l\big\|_2^2, \\
 \mathcal{L}_{\text{final}} &= \mathcal{L}_{\text{motion}} + \mathcal{L}_{\text{traj}} + \mathcal{L}_{\text{stick}}.
\end{aligned}\label{eq:samp}
\end{equation}

In this formulation, $\mathcal{L}_{\text{traj}}$ enforces global trajectory alignment between the generated motions and the ground-truth motions, ensuring spatial consistency with user-provided trajectories. $\mathcal{L}_{\text{stick}}$ regularizes pose-level fidelity by comparing predicted and ground-truth 3D poses frame by frame. A binary mask $m_i \in \{0,1\}$ is randomly sampled to allow DrawMotion to accept different combinations of stickman positions, where $M=\sum_{i=0}^L m_i$ serves as the normalization factor. Finally, $\mathcal{L}_{\text{motion}}$ constrains the reconstructed motion sequence to remain close to the reference ground-truth motion. By jointly optimizing these objectives, $\mathcal{L}_{\text{final}}$ ensures that DrawMotion learns accurate, controllable, and user-aligned motion generation.

\section{Training-free Guidance}~\label{sec:tf}

\noindent\textbf{Overview.} In DrawMotion, we further propose a novel training-free guidance termed Intermediate Feature Guidance (IFG), which is built on the Multi-Condition Module (MCM) to align the user-provided trajectory with the generated motion without additional training. The structure of this section is organized as follows: Section~\ref{sec:tfmo} introduces the motivation for such guidance and provides an overview of current works; Section~\ref{sec:space} analyzes the intermediate feature spaces of traditional models, our MCM, and generative models, explaining why the intermediate features of MCM are amenable to gradient from spatial loss; Section~\ref{sec:ma} regularizes the update process in the proposed IFG, ensuring that the fidelity of generated motions remains unaffected. 

\subsection{Motivation}~\label{sec:tfmo}
During the generation process with multiple conditions, a harmony among these conditions will eventually be reached, which may result in the generation not strictly aligning with the conditions. In particular, text control provides global semantic guidance for the motion sequences, while trajectory control provides global spatial guidance. Such conflicts may cause the trajectory of the generated motion sequence to deviate from the user-provided trajectory. To address this issue, we attempt to refer to motion editing tasks to refine the generation without compromising semantic alignment or fidelity.

As stated in Section~\ref{sec:rw}, motion editing tasks mainly focus on motion spatial guidance: The sparse spatial supervision signals in motion guidance, such as the absolute coordinates of the wrist, can be directly measured using the Euclidean distance from the generated motion. This distance is then employed as a loss function for gradient backpropagation to refine the motion. Current methods ensure the fidelity of the generated motion in two ways: 1) OmniControl~\cite{Xie2023OmniControlCA} treats $\mu_t(\epsilon_{\theta}, x_t)$ of $x_{t-1}$ in Equation~\ref{eq:reverse} as the generated motion $m$ at step $t-1$ of the diffusion reverse process, computes the loss with respect to the spatial supervision signals, and backpropagates the gradients directly to $\mu_t(\epsilon_{\theta}, x_t)$. However, this process may cause $x_{t-1}$ to deviate from its original distribution, thereby impairing fidelity. To preserve realism, ControlNet~\cite{Zhang2023AddingCC} is further employed to guide $x_{t-1}$ back to its distribution.  
2) DNO~\cite{Karunratanakul2023OptimizingDN} backpropagates the gradients of the spatial loss multiple times to the initial sampled noise $x_T$ of the diffusion process. Since $x_T$ is constrained within its prior distribution $\mathcal{N}(0, \mathbf{I})$, the perturbed noise remains consistent with this distribution. Consequently, the final generated result $x_0$ also lies within its distribution, thereby ensuring vividness. However, this approach incurs a high computational cost. Although these two approaches differ in their implementation, they share one common property: the gradients are propagated to a target variable with high tolerance, meaning that this variable follows a continuous distribution within the range of its dimensional space.

Both OmniControl and DNO bypass the direct task of maintaining the distribution of $x_{t-1}$. The former relies on ControlNet, while the latter achieves this by only perturbing $x_T$. Interestingly, we found that the intermediate features of MCM also exhibit a broad distribution, which allows us to perturb the intermediate features at step $t-1$ without causing $x_{t-1}$ to deviate from its distribution. This enables us to combine the efficiency of OmniControl with the vividness of DNO's generation without incurring additional training or computational cost. We will demonstrate this in the next subsection.

\begin{figure}[t]
\begin{center}
\includegraphics[width=1\linewidth]{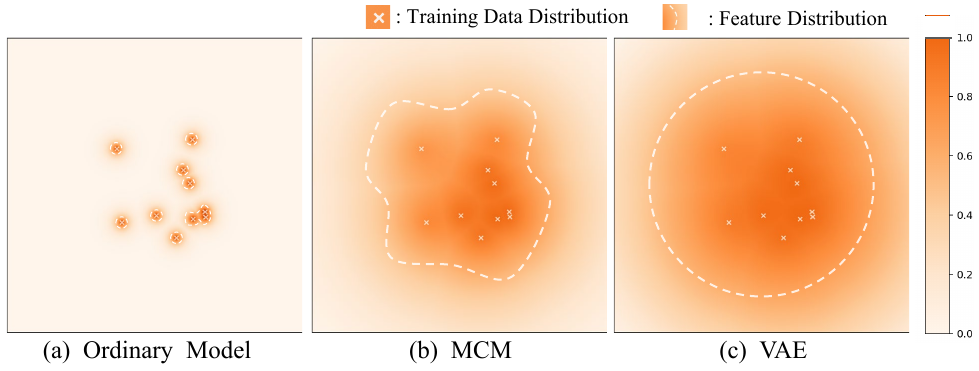}
\end{center}
\vspace{-4mm}
\caption{\small{Conceptual illustration of intermediate feature distributions. The dashed lines correspond to level sets of the probability density function. (a) Ordinary models yield discrete clusters, (b) MCM forms a relatively continuous space, and (c) VAE enforces full latent coverage. This schematic is supported by Table~\ref{tab:vae}.}}
\vspace{-2mm}
\label{fig:vae}
\end{figure}

\subsection{Intermediate Feature Space}~\label{sec:space}

The intermediate features of ordinary models are often distributed discretely in the space, \emph{i.e.}, they lie on a lower-dimensional manifold within the high-dimensional space~\cite{Tu2023ProbabilisticAS}.
For example, an AutoEncoder (AE)~\cite{hinton2006reducing} can compress images into low-dimensional intermediate features and then reconstruct them. However, it is difficult to perturb these low-dimensional features to obtain new images. This indicates that the distribution of the intermediate features in their latent space is discrete, and even slight perturbations may move them outside the distribution, as illustrated in Figure~\ref{fig:vae}(a). In contrast, VAE~\cite{kingma2013auto} addresses this issue by introducing a KL divergence loss between the intermediate features and the standard normal distribution $\mathcal{N}(0, \mathbf{I})$. This encourages the distribution of the intermediate features to cover the entire latent space, as shown in Figure~\ref{fig:vae}(c). As a result, new images can be generated from features sampled from the specified distribution $\mathcal{N}(0, \mathbf{I})$.  


\begin{figure}[h]
\begin{center}
\includegraphics[width=1\linewidth]{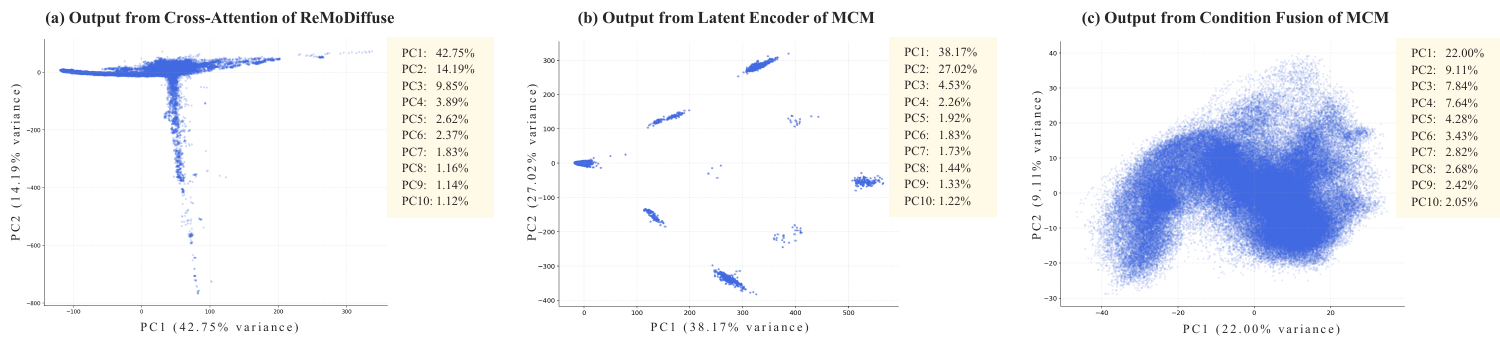}
\end{center}
\vspace{-4mm}
\caption{\small{\newm{2D PCA projection onto the first two principal components of ReMoDiffuse and DrawMotion. Sample size = 80{,}000 and diffusion step = 299.}}}
\vspace{-2mm}
\label{fig:pca}
\end{figure}

\newm{Figure~\ref{fig:pca} provides further experimental evidence supporting the above conclusion. This visualization is based on Principal Component Analysis (PCA), a linear dimensionality reduction technique that projects high-dimensional feature vectors onto orthogonal axes of maximum variance. As shown in Figure~\ref{fig:pca}: 
\textbf{(a)} The features from the last cross-attention layer of ReMoDiffuse exhibit an extremely irregular distribution.
\textbf{(b)} The output features from the Latent Encoder of the last MCM in DrawMotion, \emph{i.e.}, the features before the final linear layer, show a clustered distribution. 
\textbf{(c)} The intermediate features of MCM, namely the output features of the Condition Fusion module, show a continuous and dense distribution. This provides strong experimental support for the suitability of these intermediate features for accepting backpropagated reverse gradients. Next, we will explore the reasons for the discrete and continuous distribution of features.}

\noindent\newm{\textbf{Collapse Phenomenon of Intermediate Features in Traditional Models.} Papyan et al.~(2020)~\cite{Papyan2020PrevalenceON} empirically observed that the input features of a model's last layer within the same class collapse to their class mean in classification tasks. In other words, these features are distributed in clusters. 
Rangamani et al.~(2023)~\cite{Rangamani2023FeatureLI} extended these properties to intermediate layers through empirical studies on classification models. 
Papyan et al.~(2024)~\cite{Andriopoulos2024ThePO} theoretically proved that a collapse phenomenon also occurs at the last layer in regression tasks when weight decay regularization is used as an auxiliary loss.  Figure~\ref{fig:pca} (a) and (b)  support this conclusion. Specifically, the last-layer feature vectors collapse onto the subspace spanned by the $n$ principal components of the feature vectors, where $n$ is the dimensionality of the targets. Moreover, the same property was empirically observed even without weight decay, as shown in Appendix~A.4 of their paper, although without theoretical proof. 
Overall, these results suggest that ``\textit{the phenomenon of neural collapse could be a universal behavior in deep learning}''~\cite{Andriopoulos2024ThePO}. Next, we demonstrate through experiments that the intermediate features of MCM do not follow this rule.}

\begin{figure}[t]
\begin{center}
\includegraphics[width=1\linewidth]{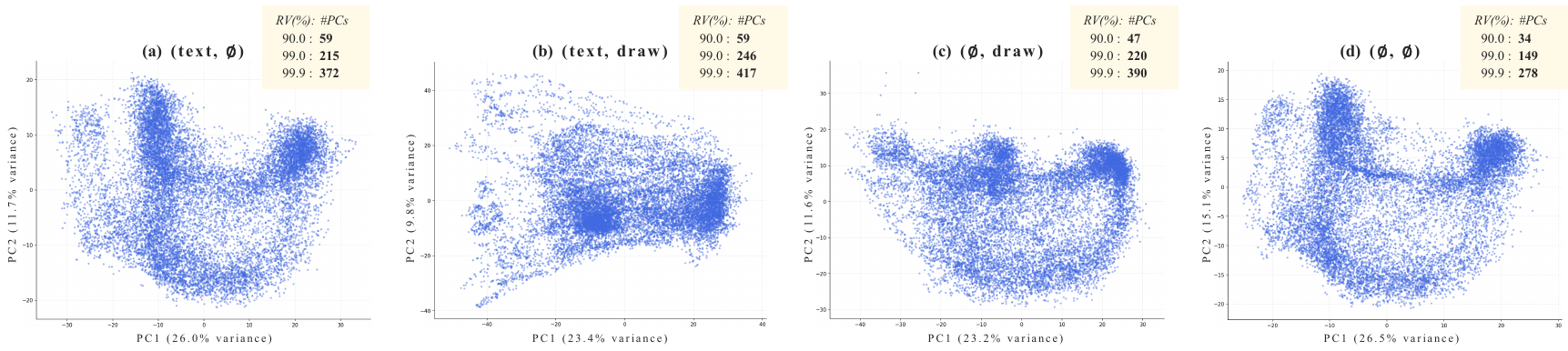}
\end{center}
\vspace{-4mm}
\caption{\small{\neww{2D PCA projection onto the first two principal components of different condition settings in DrawMotion. Sample size = 20{,}000 and diffusion step = 299.}}}
\vspace{-2mm}
\label{fig:cond_pca}
\end{figure}

\noindent\neww{\textbf{Robustness of Intermediate Features in MCM.} Although MCM does not explicitly enforce distributional alignment as done in VAEs, its intermediate features, \emph{i.e.}, the outputs of the Condition Fusion module, still exhibit a continuous and dense structure. This continuity arises from the intrinsic properties of the multi-condition fusion process. Specifically, each condition (e.g., text or drawing) is encoded into a feature representation that may \newx{lie} on a low-dimensional nonlinear manifold. The Minkowski sum of these features in the Condition Fusion module expands the effective dimensionality of the added features, leading to a higher-dimensional and more continuous feature space.}

\neww{In \newx{detail}, we separately analyze the PCA statistics for four settings: {\small$(\text{text}, \text{draw})$}, {\small$(\text{text}, \varnothing)$}, {\small$(\varnothing, \text{draw})$}, and {\small$(\varnothing, \varnothing)$}. For each case, we use the number of principal components required to explain 90\%, 99\%, and 99.9\% of the variance as a proxy for the intrinsic dimensionality of the underlying manifold. Figure~\ref{fig:cond_pca} shows that the multi-condition setting (b) exhibits a higher intrinsic dimensionality (417 dimensions for 99.9\% variance explanation) than the single-condition cases (a) and (c), and a significantly higher dimensionality than the unconditional case (d). Interestingly, the PCA visualizations reveal that the feature distributions from (d) to (a), (c), and finally to (b) may share a common geometric shape, while becoming progressively more continuous and denser. This observation suggests that (a), (c), and (d) do not reside in four independent feature spaces; instead, they can be interpreted as lower-dimensional manifold projections of (b). Consequently, we infer that the acceptable feature distribution (Figure~\ref{fig:cond_pca}.b) for  $\mathrm{Model}^{2}_{\theta}$ lies on a manifold of high intrinsic dimension. This enhances the feature's robustness to gradient-based updates, preventing minor adjustments from causing deviations from the low-dimensional manifold.}

\noindent\neww{\textbf{Empirical Evidence.}}
To further verify this inference, we conducted the following experiment. We selected ReMoDiffuse~\cite{zhang2023remodiffuse} as the baseline model for comparison. Like DrawMotion, it is a text-to-motion model with most settings kept consistent. It accepts two conditions: text and recalled reference motion. All condition combinations in ReMoDiffuse occur in a cross-attention mechanism, where the query is the input motion and the keys/values correspond to the two conditions. Unlike MCM, different condition combinations in ReMoDiffuse are achieved by masking condition inputs, which leads to unnecessary computational overhead.  

To compare their intermediate feature distributions, as illustrated in Figure~\ref{fig:vae}(a) and (b), we perturbed their intermediate features. Specifically, for DrawMotion, we selected the motion features from the Condition Fusion module (Figure~\ref{fig:main}), while for ReMoDiffuse, we selected the output of its cross-attention layer, which corresponds to our MCM, since both are used for conditional fusion.  

The next challenge is how to perturb the intermediate features. Unlike in VAEs, we do not know their exact distributions. Directly adding random noise would unfairly favor distributions with larger means, making comparison biased. Instead, we perturbed them using shuffled batches. Denote a batch of intermediate features as $F \in \mathbb{R}^{B \times E}$, where $B$ is the batch size and $E$ is the feature dimension. We then randomly shuffled the batch dimension to obtain $\hat{F} \in \mathbb{R}^{B \times E}$. The perturbed feature $\bar{F}$ is defined as:
\begin{equation} 
\bar{F} = F + \lambda (\hat{F} - F),
\label{eq:dis}
\end{equation}
where $\lambda$ is the perturbation factor.  

To examine whether the intermediate feature distribution is continuous, we interpolated the latent vectors within each batch as defined in Equation~\ref{eq:dis}. If the distribution were discrete, such interpolation would push features outside the support, leading to significant performance degradation. Indeed, this phenomenon is observed in ReMoDiffuse, as shown in Table~\ref{tab:vae}. In contrast, MCM maintains stable generation quality across a wide range of interpolation factors, indicating that its feature distribution forms connected regions rather than isolated atoms. This provides strong empirical evidence that MCM learns a relatively continuous latent space~\cite{kingma2013auto}.  

Finally, we demonstrate that the intermediate features of MCM can tolerate gradient-based perturbations, which allows us to perturb the intermediate features at step $t-1$ without causing $x_{t-1}$ to deviate from its distribution. Therefore, the intermediate features can be updated directly, without requiring additional modules such as ControlNet.  

\begin{table}[t]
\centering
\caption{\small{Comparison of FID under different perturbation factors $\lambda$. Lower is better.}}
\small
\renewcommand\arraystretch{1.2}
\setlength{\tabcolsep}{3.5mm}
\begin{tabular}{ccc}
\toprule
 $\lambda$ & FID$\downarrow$ {\footnotesize ReMoDiffuse} & FID$\downarrow$ {\footnotesize DrawMotion} \\ 
\midrule
 \rowcolor{mygray} 0\% & 0.159 & 0.146 \\
 1\% & 0.283 & 0.143 \\
 \rowcolor{mygray} 10\% & 29.67 & 0.141 \\
 30\% & 73.15 & 0.143 \\
 \rowcolor{mygray} 50\% & 117.3 & 0.171 \\
\bottomrule
\end{tabular}\label{tab:vae}
\vspace{-3mm}
\end{table}

\subsection{Intermediate Feature Guidance}~\label{sec:ma}

Based on the above analysis, we can update the intermediate features $F$ using Stochastic Gradient Descent (SGD) to meet the spatial signal requirements. Although we do not know the exact distribution of $F$ as shown in Figure~\ref{fig:vae}, we can ensure that the updated $\hat{F}$ does not deviate too much from the distribution by increasing the number of SGD iterations and reducing the learning rate of the update, as shown in rows 1-5 of Table~\ref{tab:ma}. If we want to accelerate this SGD process, we must increase the learning rate, which introduces uncertainty about whether the gradient $\nabla_{\bar{F}}$ of this iteration will increase or decrease $p_\theta(\bar{F})$. We introduce the Mahalanobis distance~\cite{Mahalanobis1936OnTG} to solve this problem, as shown in Algorithm~\ref{alg:alg1}.

\definecolor{algc}{rgb}{0.2, 0.6, 0.3}

\begin{algorithm}[h]
\caption{Intermediate Feature Guidance}\label{alg:alg1}
$\mathbf{x}_T \sim \mathcal{N}(0, \mathbf{I})$\;
{\color{algc}\textit{// Reverse process.}}\\

\For{$t = T, T-20, T-40, \ldots, 1$}{
    {\color{algc}\textit{// Extract the intermediate feature $F$.}}\\
    
    $F \gets \mathrm{Model}^1_\theta(\mathbf{x}_t,t,L,C(\text{draw}), C(\text{text}))$\;
    $\bar{F} \gets F$\;
    {\color{algc}\textit{// Update $F$ with SGD.}}\\
    
    \For{$i \gets 1, \ldots, R$}{
        {\color{algc}\textit{// Get the predicted $\hat{x_0}$ from the predicted noise $\epsilon_\theta$. Here define $f(\cdot)$ cf. Eq.~\ref{eq:ddim}.}}\\
        $\hat{x}_{0}(\bar{F}, \dots) \gets f(\mathrm{Model}^2_\theta(F, \dots))$; 
        
        {\color{algc}\textit{// Update the intermediate feature $\bar{F}$.}}\\
        $G_{\bar{F}} \gets \bar{F} - lr \cdot \nabla_{\bar{F}}||\hat{x}_{0}(\bar{F}, \dots) - c||_2^2$\;
        
        {\color{algc}\textit{// Let $M(F)$ denotes the \textbf{Mahalanobis distance} between $F$ and the statistical distribution.}}\\
        
        \If{$M(\bar{F}) > M(F) + \epsilon^{MD}$}{
            {\color{algc}\textit{// MD clipping.}}\;
            $\bar{F} \gets F + \lambda\times(\bar{F} - F)$\;
        }
    }
    {\color{algc}\textit{// Define $g(\cdot)$ cf. Eq.~\ref{eq:ddim}.}}\\
    
    $\mathbf{x}_{t-1} \gets g(\epsilon^2_\theta(F, \dots))$;
}
\Return{$\mathbf{x}_0$}\\
\end{algorithm}
Here, we assume that the intermediate feature $F$ of the $N_{th}$ MCM layer is used as the optimization objective, and the model is divided into two parts with respect to this feature, which are noted as $\mathrm{Model}^1$ and $\mathrm{Model}^2$ respectively. During the DDIM reverse process, we first obtain the intermediate feature $F$ from $\mathrm{Model}^1$, then we get $\bar{F}$ closer to the $F^{optimal}$ for \newm{guidance} loss $||\hat{x}_{0}(\bar{F}, \dots) - c||_2^2$ through SGD. Here, $c$ is the spatial guidance with the same shape as $\hat{x_t}$, and the unconstrained parts with no spatial guidance do not participate in the loss calculation by the mask method. During SGD, MD Clipping is proposed to ensure that the updated $\bar{F}$ does not deviate from the statistical distribution with the clip scale $\lambda$ (details are provided below). Finally, based on $\bar{F}$, we can obtain an $\mathbf{x}_{t-1}$ that is both of high fidelity and guided by spatial signals, and then continue the reverse process until we obtain the target $x_0$. 

\noindent\textbf{MD Clipping.} 
To constrain the intermediate feature $\bar{F}$ within a plausible region of the high-dimensional feature space, we leverage the Mahalanobis Distance (MD), which measures the deviation of a sample from a multivariate distribution while accounting for correlations between features. Specifically, we estimate the mean $\mu$ and covariance $\Sigma$ of the intermediate features during evaluation and define $M(F) = \sqrt{(F-\mu)^T \Sigma^{-1} (F-\mu)}$. During the SGD-based update of $\bar{F}$, we monitor $M(\bar{F})$ and perform gradient clipping whenever the Mahalanobis distance reaches the MD boundary $M(F) + \epsilon^{MD}$, which is the sum of the origin distance and a threshold. This method is called MD clipping, which ensures that updates driven by the reconstruction loss $||\hat{x}_{0}(\bar{F},\dots) - c||_2^2$ remain within the high-probability region of the feature distribution, preventing out-of-distribution artifacts. Unlike Euclidean distance, Mahalanobis distance adapts to feature variance and correlations, offering a statistical constraint suitable for high-dimensional latent spaces. Empirically, this approach stabilizes the spatial guidance in the reverse diffusion process, improves the performance of DrawMotion (Table~\ref{tab:ma}), and shortens the inference time (Table~\ref{tab:exp_traj}).

\begin{table}[t]
\centering
\caption{\small{Hyperparameter analysis of Intermediate Feature Guidance (IFG) on KIT-ML dataset. 
Here, $repeat$ denotes the number of SGD iterations, 
$lr$ is the learning rate of this update, 
$N_{th}$ layer specifies the selected MCM layer for guidance, 
$\epsilon^{MD}$ is the Mahalanobis distance threshold used for clipping abnormal updates,
and $\lambda$ is the clip scale. }}
\small
\renewcommand\arraystretch{1.2}
\setlength{\tabcolsep}{1.7mm}
\begin{tabular}{c ccccccc}
\toprule
 & $repeat$ & $lr$ & $N_{th}$ layer & $\epsilon^{MD}$ & $\lambda$ & Traj.err.$\downarrow$ & FID$\downarrow$ \\
\toprule
1  & 100 & 10 & 1 & N/A  & N/A & \cellcolor{blue!32}0.112 & \cellcolor{orange!58}0.131 \\
2  & 100 & 10 & 2 & N/A  & N/A & \cellcolor{blue!35}0.105 & \cellcolor{orange!52}0.139 \\
3  & 100 & 10 & 3 & N/A  & N/A & \cellcolor{blue!39}0.099 & \cellcolor{orange!48}0.146 \\
\hline
4  & 50  & 20 & 3 & N/A  & 0.5 & \cellcolor{blue!30}0.114 & \cellcolor{orange!40}0.163 \\
5  & 10  & 50 & 3 & N/A  & 0.5 & \cellcolor{blue!20}0.126 & \cellcolor{orange!30}0.185 \\
\hline
6  & 10  & 50 & 3 & -10  & 0.5 & \cellcolor{blue!20}0.126 & \cellcolor{orange!52}0.141 \\
7  & 10  & 50 & 3 & -5   & 0.5 & \cellcolor{blue!20}0.126 & \cellcolor{orange!54}0.140 \\
8  & 10  & 50 & 3 & -1   & 0.5 & \cellcolor{blue!22}0.125 & \cellcolor{orange!54}0.140 \\
9  & 10  & 50 & 3 & 0    & 0.5 & \cellcolor{blue!34}0.096 & \cellcolor{orange!52}0.141 \\
10 & 10  & 50 & 3 & 1    & 0.5 & \cellcolor{blue!40}0.069 & \cellcolor{orange!52}0.141 \\
11 & 10  & 50 & 3 & 10   & 0.5 & \cellcolor{blue!37}0.084 & \cellcolor{orange!52}0.141 \\
12 & 10  & 50 & 3 & 50   & 0.5 & \cellcolor{blue!31}0.102 & \cellcolor{orange!36}0.167 \\
\hline
13 & 10  & 50 & 3 & 1    & 0.5 & \cellcolor{blue!40}0.069 & \cellcolor{orange!52}0.141 \\
14 & 10  & 50 & 3 & 1    & 0.3 & \cellcolor{blue!41}0.065 & \cellcolor{orange!57}0.137 \\
15 & 10  & 50 & 3 & 1    & 0.1 & \cellcolor{blue!42}0.062 & \cellcolor{orange!57}0.137 \\
16 & 10  & 50 & 3 & 1    & 0.01 & \cellcolor{blue!43}0.061 & \cellcolor{orange!60}0.135 \\
17 & 10  & 50 & 3 & 1    & 0.001 & \cellcolor{blue!43}0.061 & \cellcolor{orange!56}0.138 \\
18 & 10  & 50 & 3 & 1    & 0.0 & \cellcolor{blue!43}0.061 & \cellcolor{orange!59}0.136 \\
19 & 10  & 50 & 3 & 1    & -0.1 & \cellcolor{blue!44}0.060 & \cellcolor{orange!55}0.139 \\
\hline
20 & 50  & 50 & 3 & 1    & 0.01 & \cellcolor{blue!48}0.032 & \cellcolor{orange!65}0.135 \\
21 & 100 & 50 & 3 & 1    & 0.01 & \cellcolor{blue!50}0.026 & \cellcolor{orange!70}0.132 \\
\bottomrule
\end{tabular}
\label{tab:ma}
\vspace{-3mm}
\end{table}

\noindent\textbf{Hyperparameter Tuning.}  
We analyze the effect of different hyperparameters in Table~\ref{tab:ma}:  
1) \textit{Layer selection (rows 1--3).} Deeper layers (closer to the output) lead to lower trajectory error (Traj.err.) but higher FID. We therefore select $N_{th}=3$, which strikes a balance while also reducing computation.  
2) \textit{Repeat count (rows 4--5).} $repeat$ denotes the number of SGD iterations applied to $\bar{F}$ under spatial guidance, with $lr$ adjusted accordingly. Fewer iterations require a larger $lr$, which increases the risk of drifting outside the valid feature distribution and results in degraded generation quality.  
3) \textit{MD threshold $\bm{\epsilon^{MD}}$ (rows 6--12).} Enabling Mahalanobis distance (MD) clipping yields substantial improvements in both Traj.err. and FID, especially at $\epsilon^{MD}=1$. When $\epsilon^{MD} < 1$, gradients are consistently clipped, preventing $F$ from moving and thus harming Traj.err., though FID remains stable. Conversely, overly large thresholds cause both metrics to deteriorate, as the updated $\mathbf{x}_{t-1}$ deviates too far from the distribution and is treated as noise in subsequent reverse steps.  
4) \textit{Clip scale $\lambda$ (rows 13--19).} $\lambda$ determines how updates behave when $\bar{F}$ reaches the MD boundary $M(F)+\epsilon^{MD}$. The results suggest that the best practice is to retain only a very small portion (around 0.01) of the update gradient $\bar{F}-F$ once the boundary is exceeded, while discarding the rest. This effectively prevents destabilization while still providing minimal perturbations that help $F$ explore new directions---a behavior that is also theoretically justified.  
5) \textit{Best practice (rows 20--21).} Among all tested settings, row 16 provides a good trade-off between Traj.err. and FID. Increasing the repeat count under this configuration further improves results, though at the expense of higher computation. In the following experiments (Section~\ref{sec:exp}), we adopt the configuration from row 16.

\begin{table*}[t]
\renewcommand\arraystretch{1.3}
\setlength{\tabcolsep}{1.6mm}
\caption{\small{Comparison on the HumanML3D test set. We mark the best result as \colorbox{red!15}{red} and the second best one as \colorbox{blue!15}{blue}. Arrows indicate the desired direction of metrics: $\downarrow$ (lower is better), $\uparrow$ (higher is better), and $\to$ (closer to real data is better). }}
\vspace{-2mm}
\centering
\begin{tabular}{lcccccccc}
\hline 
\multirow{2}{*}{Methods} & \multirow{2}{*}{FID $\downarrow$} & \multicolumn{3}{c}{R Precision $\uparrow$} & \multirow{2}{*}{MM Dist $\downarrow$} & \multirow{2}{*}{Diversity  $\to$} & \multirow{2}{*}{MultiModality $\uparrow$} & \multirow{2}{*}{StiSim $\uparrow$} \\ 
&  & Top1 & Top2 & Top3 & & & & \\
\hline 
Real motions &  0.002$^{ \pm .000}$  &  0.511$^{ \pm .003}$  &  0.703$^{ \pm .003}$  &  0.797$^{ \pm .002}$  &  2.974$^{ \pm .008}$  &  9.503$^{ \pm .065}$  & - & - \\
\hline 
Guo et al.~\cite{guo2022generating} &  1.067$^{ \pm .002}$  &  0.457$^{ \pm .002}$  &  0.639$^{ \pm .003}$  &  0.740$^{ \pm .003}$  &  3.340$^{ \pm .008}$  &  9.188$^{ \pm .002}$  &  2.090$^{ \pm .083}$  & - \\
MDM~\cite{tevet2022humanmotiondiffusionmodel} &  0.544$^{ \pm .044}$  & - & - &  0.611$^{ \pm .007}$  &  5.566$^{ \pm .027}$  &\cellcolor{blue!15}   9.559$^{ \pm .086}$  &\cellcolor{red!15}   2.799$^{ \pm .072}$  & - \\
MotionDiffuse~\cite{zhang2022motiondiffuse} &  0.630$^{ \pm .001}$  &  0.491$^{ \pm .001}$  &  0.681$^{ \pm .001}$  &  0.782$^{ \pm .001}$  &  3.113$^{ \pm .001}$  &  9.410$^{ \pm .049}$  &  1.553$^{ \pm .042}$  & - \\
T2M-GPT~\cite{zhang2023generating} &  0.116$^{ \pm .004}$  &  0.491$^{ \pm .003}$  &  0.680$^{ \pm .003}$  &  0.775$^{ \pm .002}$  &  3.118$^{ \pm .011}$  &   9.761$^{ \pm .081}$  &  1.856$^{ \pm .011}$  & - \\
ReMoDiffuse~\cite{zhang2023remodiffuse} &\cellcolor{red!15}   0.103$^{ \pm .004}$  &\cellcolor{blue!15}   0.510$^{ \pm .005}$  &\cellcolor{blue!15} 0.698$^{ \pm .006}$  &\cellcolor{blue!15}  0.795$^{ \pm .004}$  &\cellcolor{blue!15}  2.974$^{ \pm .016}$  &  9.018$^{ \pm .075}$  &  1.795$^{ \pm .043}$  & - \\\hline
StickMotion (Ours)~\cite{wang2025stickmotion} &\cellcolor{blue!15}   0.107$^{ \pm .003}$  &\cellcolor{red!15}   0.518$^{ \pm .007}$  &\cellcolor{red!15}    0.702$^{ \pm .003}$  &\cellcolor{red!15}   0.797$^{ \pm .005}$  &\cellcolor{red!15}   2.953$^{ \pm .021}$  &  9.239$^{ \pm .066}$  &  \cellcolor{blue!15} 2.256$^{ \pm .051}$ & \cellcolor{blue!15} 41.50\% \\

DrawMotion (Ours) &  0.108$^{ \pm .004}$  &   0.504$^{ \pm .004}$  &   0.695$^{ \pm .004}$  &   0.792$^{ \pm .004}$  &   2.992$^{ \pm .020}$  & \cellcolor{red!15} 9.553$^{ \pm .069}$  &  1.241$^{ \pm .057}$ &\cellcolor{red!15}  59.26\% \\
\hline
\end{tabular}\label{tab:t2m}
\end{table*}

\begin{table*}[t]
\renewcommand\arraystretch{1.3}
\setlength{\tabcolsep}{1.6mm}
\caption{\small{Comparison on the KIT-ML test set. }}
\vspace{-2mm}
\centering
\begin{tabular}{lcccccccc}
\hline 
\multirow{2}{*}{Methods} & \multirow{2}{*}{FID $\downarrow$} & \multicolumn{3}{c}{R Precision $\uparrow$} & \multirow{2}{*}{MM Dist $\downarrow$} & \multirow{2}{*}{Diversity $\to$} & \multirow{2}{*}{MultiModality $\uparrow$} & \multirow{2}{*}{StiSim $\uparrow$} \\ 
&  & Top1 & Top2 & Top3 & & & & \\
\hline 
Real motions &  0.031$^{ \pm .004}$  &  0.424$^{ \pm .005}$  &  0.649$^{ \pm .006}$  &  0.779$^{ \pm .006}$  &  2.788$^{ \pm .012}$  &  11.08$^{ \pm .097}$  & - & -\\
\hline 
Guo et al.~\cite{guo2022generating} &  2.770$^{ \pm .109}$  &  0.370$^{ \pm .005}$  &  0.569$^{ \pm .007}$  &  0.693$^{ \pm .007}$  &  3.401$^{ \pm .008}$  &  10.91$^{ \pm .119}$  &  1.482$^{ \pm .065}$  & - \\
MDM~\cite{tevet2022humanmotiondiffusionmodel} &  0.497$^{ \pm .021}$  & - & - &  0.396$^{ \pm .004}$  &  9.191$^{ \pm .022}$  &  10.85$^{ \pm .109}$  & \cellcolor{red!15} 1.907$^{ \pm .214}$  & - \\
MotionDiffuse~\cite{zhang2022motiondiffuse} &  1.954$^{ \pm .062}$  &  0.417$^{ \pm .004}$  &  0.621$^{ \pm .004}$  &  0.739$^{ \pm .004}$  &  2.958$^{ \pm .005}$  & \cellcolor{red!15} 11.10$^{ \pm .143}$  &  0.730$^{ \pm .013}$  & - \\
T2M-GPT~\cite{zhang2023generating} &  0.514$^{ \pm .029}$  &  0.416$^{ \pm .006}$  &  0.627$^{ \pm .006}$  &  0.745$^{ \pm .006}$  &  3.007$^{ \pm .023}$  &  10.92$^{ \pm .108}$  & \cellcolor{blue!15} 1.570$^{ \pm .039}$  & - \\
ReMoDiffuse~\cite{zhang2023remodiffuse} & 0.155$^{ \pm .006}$  & \cellcolor{blue!15} 0.427$^{ \pm .014}$  &  0.641$^{ \pm .004}$  & 0.765$^{ \pm .055}$  & 2.814$^{ \pm .012}$  &  10.80$^{ \pm .105}$  &  1.239$^{ \pm .028}$  & - \\\hline
StickMotion (Ours)~\cite{wang2025stickmotion} &\cellcolor{blue!15} 0.141$^{ \pm .008}$  &\cellcolor{red!15} 0.430$^{ \pm .017}$  & \cellcolor{red!15}  0.654$^{ \pm .010}$  & \cellcolor{red!15} 0.775$^{ \pm .043}$  & \cellcolor{red!15} 2.763$^{ \pm .018}$  & 10.94$^{ \pm .178}$  &  1.457$^{ \pm .033}$ & \cellcolor{blue!15} 42.60\% \\ 

DrawMotion (Ours) &\cellcolor{red!15} 0.135$^{ \pm .007}$  & 0.423$^{ \pm .010}$  & \cellcolor{blue!15}  0.643$^{ \pm .007}$  &  \cellcolor{blue!15} 0.776$^{ \pm .006}$  &  \cellcolor{blue!15}  2.772$^{ \pm .003}$  & \cellcolor{blue!15} 10.92$^{ \pm .130}$  &  0.916$^{ \pm .006}$ &\cellcolor{red!15}  52.17\% \\
\hline
\end{tabular}\label{tab:kit}
\end{table*}

\section{Experiments}~\label{sec:exp}

\subsection{Experiment Settings}
   
\noindent\textbf{Dataset and Metrics.}\label{metric}
We conduct experiments on two popular datasets of human motion generation, namely the KIT-ML dataset~\cite{plappert2016kit} and the HumanML3D dataset~\cite{guo2022generating}. \neww{The motion representation consists of local skeleton poses relative to the root and global root translations and rotations across frames.} The same evaluation protocol as Guo et al.~\cite{guo2022generating} is adopted, so we can comprehensively compare with existing text-to-motion methods. This evaluation involves encoding the input text and generated motion sequence into embeddings through pre-trained contrastive quantitative assessment models, and then calculating the following metrics:
\textit{R Precision.} Given a predicted motion sequence, its text and 31 other irrelevant texts from the test set are combined into a set, and the Top-k accuracy between the motion and text set is calculated after passing through the pre-trained motion-text contrastive models. 
\textit{Frechet Inception Distance (FID).} Motion features are generated from ground-truth and generated motion sequences through contrastive models. Then, the difference between the two batches of embedding distributions is calculated. FID is related to generation quality but is limited by the performance of contrastive models.
\textit{Multimodal Distance (MM Dist).} The Euclidean distance between the motion feature and its text embedding.
\textit{Diversity.} The variability and richness of the generated motion sequences.
\textit{Multimodality.} The variance of generated motion sequences given a specified text.

Moreover, the \textit{Stickman Similarity (StiSim)}~\cite{wang2025stickmotion} and 2D \textit{Trajectory error (Traj. err.)}~\cite{Xie2023OmniControlCA} between the generated motion sequences and the given trajectories are also reported for comparison with existing motion editing tasks.

\noindent\textbf{Implementation Details.}
DrawMotion is trained with 4 4090 GPUs and a batch size of 1024, while 40 dataloader workers are used to generate stickmen through SGA. The trajectories used for input are directly obtained from the corresponding motion sequences. For the diffusion process, we set the noise steps $T=1000$. Additionally, $\alpha_t$, where $t\in [0,T]$, ranges from 0.9999 to 0.9800. The trainable DrawMotion model comprises 4 MCM layers for the KIT-ML dataset and 6 MCM layers for the HumanML3D dataset, with parameter counts of 208M and 227M respectively (including condition encoders).

\subsection{Quantitative Analysis}

\noindent\textbf{Comparison with SOTA Methods.} We follow the same evaluation protocol of text-to-motion methods to demonstrate the performance of DrawMotion based on both the KIT-ML dataset and the HumanML3D dataset. Additionally, the generation approach of the input stickman and trajectory used in the evaluation is similar to that of the training process as described in Section~\ref{sec:loss}, and the selection of stickmen is the same as StickMotion~\cite{wang2025stickmotion}, that is, the beginning, middle, and end of the motion sequence. In contrast to conventional approaches, DrawMotion requires alignment with both textual descriptions and drawings. As mentioned in Section~\ref{sec:abl_mixture}, these two conditions have an adversarial relationship that can negatively impact DrawMotion's performance. Hence, we adjusted the reverse process mentioned in Section~\ref{sec:diffusion} by setting $p(\hat{w}= w)=20\%$ and $(w_1=1,w_2=0,w_3=0,w_4=0)$ to bias the generated results towards the hand-drawing condition. Our approach shows excellent performance compared to previous text-to-motion works as shown in Table~\ref{tab:t2m} and Table~\ref{tab:kit}.

\noindent\textbf{Stickman Similarity (StiSim).}
Compared with StickMotion~\cite{wang2025stickmotion}, DrawMotion has a higher StiSim as shown in Table~\ref{tab:t2m} and \ref{tab:kit} due to the following reasons: 1) In DrawMotion, the position of the stickman in the motion sequences is explicitly specified, whereas StickMotion must determine this position internally, which may lead to ambiguity. 2) StickMotion utilizes at most 3 stickmen per motion sequence, while DrawMotion employs an average of 7 during training, thereby significantly increasing the amount of training data. 3) The cross-attention structure we adopted for stickman conditions is better suited to the task than the one used in StickMotion, as demonstrated in Section~\ref{sec:network}.

\begin{table}[t]
\centering
\setlength{\tabcolsep}{2mm}
\caption{Comparison of diffusion-based motion edit methods on T2M and KIT datasets. R-prec (top3) denotes R-precision (top3).}
\begin{tabular}{c c c c c}
\toprule
Dataset & Method & FID $\downarrow$ &  R-prec (top3) $\uparrow$ & Traj.Err. $\downarrow$ \\
\midrule
\multirow{6}{*}{\makecell{Human\\ML3D}} & 
MDM~\cite{tevet2022humanmotiondiffusionmodel} & 0.698 & 0.602 & 0.8131 \\
& GMD~\cite{Karunratanakul2023GuidedMD} & 0.576 & 0.665 & 0.6892 \\
& \neww{PriorMDM~\cite{Shafir2023HumanMD}} &0.475 & 0.583 & 0.7412 \\
& \neww{CondMDI~\cite{Cohan2024FlexibleMI}} & 0.247 & 0.675 & 0.1178 \\
& DNO~\cite{Karunratanakul2023OptimizingDN} & 2.464 & 0.522 & 0.1057 \\
& OmniControl~\cite{Xie2023OmniControlCA} & 0.218 & 0.687 & 0.0664 \\
& DrawMotion (Ours) & \textbf{0.108} & \textbf{0.792} & \textbf{0.0062} \\
\midrule
\multirow{3}{*}{\makecell{KIT\\-ML}} & 
\neww{PriorMDM~\cite{Shafir2023HumanMD}} & 0.851 & 0.397 & 0.627 \\
& OmniControl~\cite{Xie2023OmniControlCA} & 0.702 & 0.397 & 0.238 \\
& DrawMotion (Ours) & \textbf{0.135} & \textbf{0.776} & \textbf{0.032} \\
\bottomrule
\end{tabular}
\label{tab:exp_traj}
\end{table}

\begin{table}[h]
\renewcommand\arraystretch{1.2}
\setlength{\tabcolsep}{1.3mm}
\caption{\small{Comparison of efficiency on the HumanML3D dataset with a batch size of 16. The number in Method indicates the step of the diffusion reverse process.}}\label{tab:speed}
\vspace{-2mm}
\centering
\begin{tabular}{lccc}
\hline
Name & GPU Memory (MB) & Time/Batch (s) & Method \\
\hline
OmniControl~\cite{Xie2023OmniControlCA}  & 2,145   & 153 & DDPM-1000 \\
DNO~\cite{Karunratanakul2023OptimizingDN} & 22,727  & 358 & DDIM-10   \\
DrawMotion (Ours)  & 2,245   & 24  & DDIM-50   \\
\hline
\end{tabular}\label{tab:time_traj}
\end{table}

\noindent\textbf{Motion Edit.}
As shown in Table~\ref{tab:exp_traj} and Table~\ref{tab:time_traj}, we compared DrawMotion with other diffusion-based motion editing methods. It is obvious that DrawMotion achieves the best performance and speed. Theoretically, DNO~\cite{Karunratanakul2023OptimizingDN} should have a better FID. However, in practical applications, due to GPU memory limitations, the official implementation used the diffusion process of DDIM-10. 
\neww{All training-free motion editing methods~\cite{tevet2022humanmotiondiffusionmodel, Karunratanakul2023OptimizingDN} exhibit poor performance in terms of FID, since the model cannot effectively handle data that deviate from the training distribution. In contrast, purely training-based methods~\cite{Shafir2023HumanMD, Cohan2024FlexibleMI} achieve only moderate trajectory error due to the lack of additional constraints.}


\subsection{Ablation Study}

\noindent\textbf{Analysis on the Structure of Condition Decoders}\label{sec:abl_decoder}
We first perform ablation studies on the structure of condition decoders using the KIT-ML dataset, as shown in Table~\ref{tab:abl_decoder}. As discussed in Section~\ref{sec:network}, two types of structures are considered: dot-product attention and efficient attention. Note that the dot-product operation corresponds to the standard self-attention mechanism. Based on their underlying computational logic, we argue that text conditions, which are more global in nature, are better suited for efficient attention, whereas drawing conditions, which emphasize local details, benefit more from dot-product attention. The results in Table~\ref{tab:abl_decoder} support this observation.

\begin{table}[t]
\centering
\setlength{\tabcolsep}{1.9mm}
\caption{\small{Analysis on the Structure of Condition Decoders  on the KIT-ML dataset. R-prec (top3) denotes R-precision (top3). Text/Draw denotes the Text/Draw Decoder. And \emph{dot}/\emph{eff} denotes the dot-product/efficient attention structure respectively. The row with a gray background is our best practice.}}
\begin{tabular}{c c |c c c c}
\toprule
Text & Draw & FID $\downarrow$ & R-prec (top3) $\uparrow$ & StiSim $\uparrow$ & Traj.Err. $\downarrow$ \\
\midrule
\emph{dot} & \emph{dot} & 0.153 & 0.729 & \textbf{52.3\%} & 0.041 \\
\rowcolor{mygray} \emph{eff} & \emph{dot} & \textbf{0.135} & \textbf{0.776} & 52.2\% & \textbf{0.032} \\
\emph{dot} & \emph{eff} & 0.147 & 0.742 & 46.5\% & 0.085 \\
\emph{eff} & \emph{eff} & 0.141 & 0.768 & 45.7\% & 0.113 \\
\bottomrule
\end{tabular}\label{tab:abl_decoder}
\end{table}

\begin{table}[!h]
\centering
\setlength{\tabcolsep}{1.3mm}
\caption{\small{Analysis on the Structure of MCM  on the KIT-ML dataset. Rows without $\surd$ in the column ``Condition Fusion" mean use of the traditional mask mechanism. Rows without $\surd$ in the column "Latent Encoder" mean a simple linear layer is used. The row with a gray background is our best practice.}} 
\begin{tabular}{c c |c c c c c}
\toprule
\makecell{Condition\\Fusion} & \makecell{Latent\\Encoder} & FID $\downarrow$ & \makecell{R-prec\\(top3)} $\uparrow$ & StiSim $\uparrow$ & Traj.Err. $\downarrow$  & TFlops $\downarrow$ \\
\midrule
 &  & 0.151 & 0.764 & 50.6\% & 0.048 &0.46\\
 $\surd$&  & 0.187 & 0.759 & 51.5\% & 0.063 &\textbf{0.28} \\
 & $\surd$& 0.143 & 0.767 & 51.0\% & 0.044 &0.71 \\
\rowcolor{mygray} $\surd$ & $\surd$ &  \textbf{0.135} & \textbf{0.776} & \textbf{52.2\%} & \textbf{0.032} &0.43\\
\bottomrule
\end{tabular}\label{tab:abl_mcm}
\end{table}

\noindent\textbf{Analysis on the Structure of Multi-Condition Module.}\label{sec:abl_mcm} After we determined the structure of decoders, we further conducted ablation experiments on the MCM structure. As shown in Table~\ref{tab:abl_mcm}, we replaced Condition Fusion and the Latent Encoder with the traditional mask mechanism~\cite{zhang2023remodiffuse} and a fully connected layer, respectively, to test the validity of both components in the MCM. The first row of Table~\ref{tab:abl_mcm} represents the implementation of the traditional mask method, and the second row achieves the minimum computational demand. However, an additional latent encoder will further enhance the model's performance.

\begin{figure*}[]
\begin{center}
\includegraphics[width=0.98\linewidth]{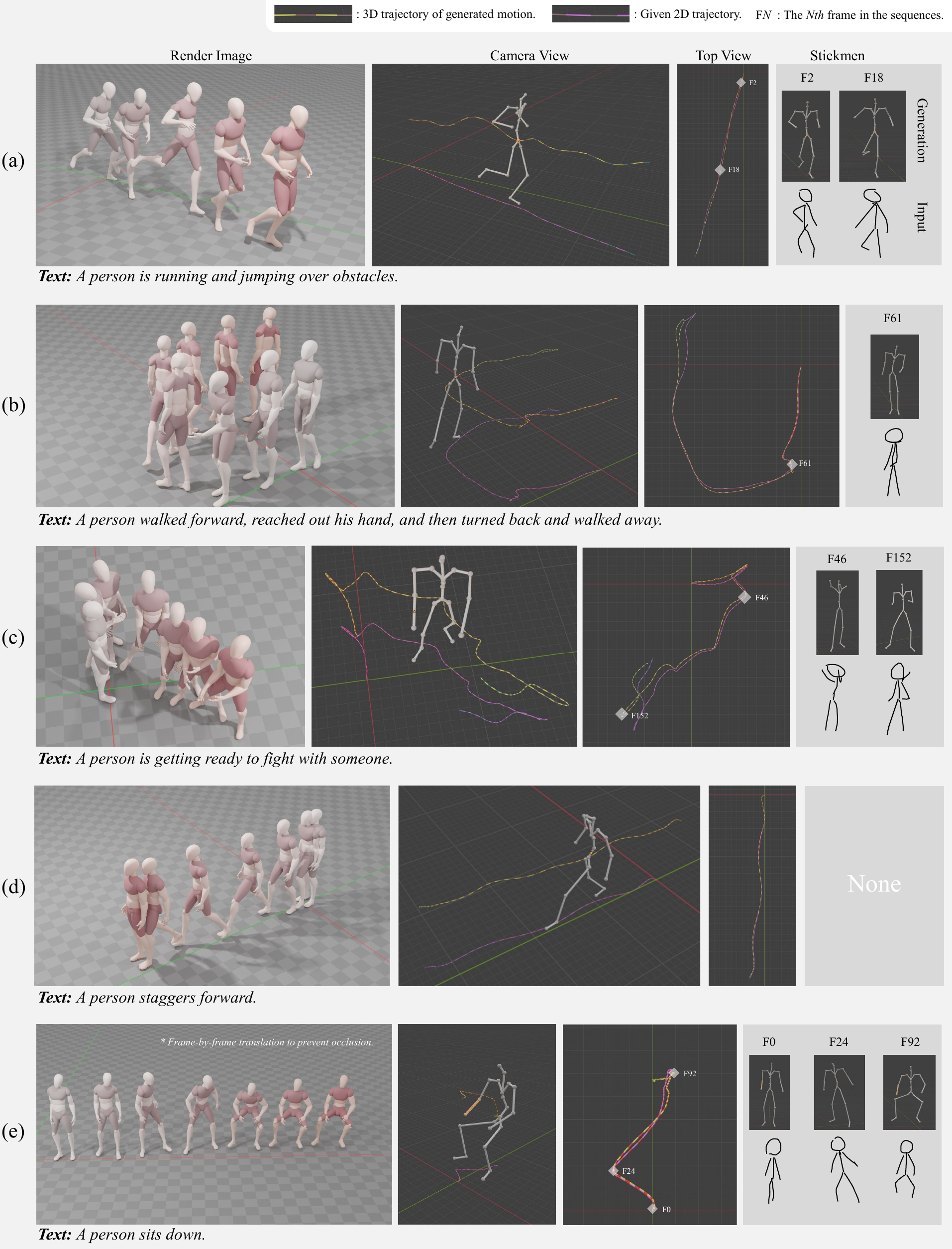}
\end{center}
\caption{\small{Visualization of DrawMotion (see the animation on GitHub). }}
\label{fig:vis}
\end{figure*}

\begin{figure}[t]
\begin{center}
\includegraphics[width=1.0\linewidth]{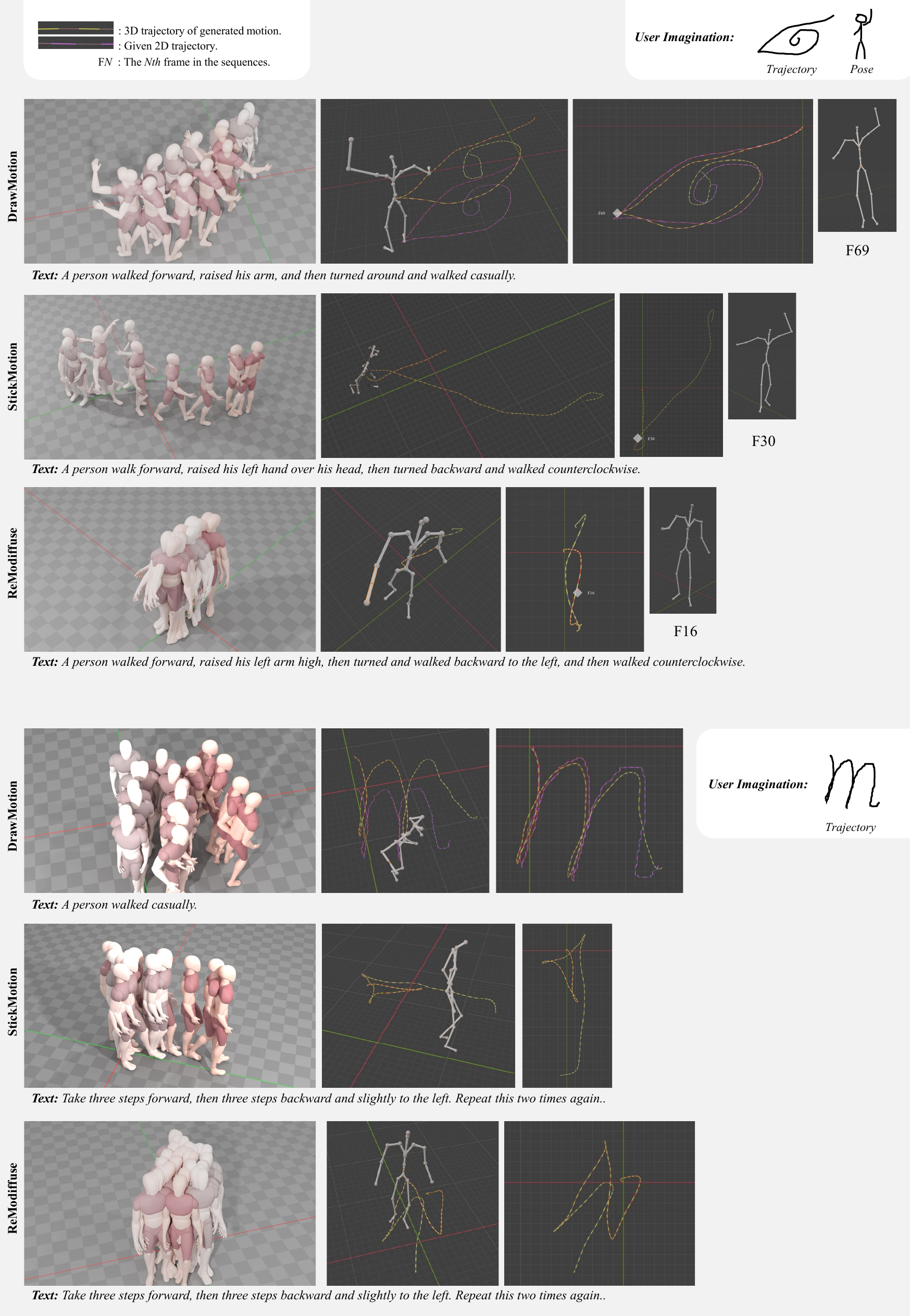}
\end{center}
\caption{\small{Visual comparison between ReModiffuse, StickMotion, and DrawMotion: 1) This user attempted to make the generated trajectory resemble the emblem from Naruto and specified that, at a designated position along the trajectory, the action should involve raising the left hand high. 2) This user simply wrote the letter "m", without specifying a stickman. (see the animation on GitHub).}}
\label{fig:compare}
\end{figure}

\begin{table}[t]
\centering
\caption{\small{Ablation study on the condition mixture for the inference / reverse process on KIT-ML dataset. The row with a gray background is our best practice.}}
\small
\renewcommand\arraystretch{1.2}
\setlength{\tabcolsep}{2.3mm}
\begin{tabular}{cccccccc}
\toprule
 $p(\hat{w}= w)$ & $(w_1,w_2,w_3,w_4)$ & FID$\downarrow$ & StiSim$\uparrow$  & Traj.err.$\downarrow$    \\ \toprule
50\%   &   (0, 1, 0, 0)  &  \textbf{0.124}  & 50.0\%  & 0.031 \\
50\%   &   (0, 0, 1, 0)   &  0.142   & 53.8\% & \textbf{0.030} \\
\rowcolor{mygray} 20\%    & (1, 0, 0, 0)  & 0.135     &    52.2\% & 0.032 \\
80\%  &  (1, 0, 0, 0)  &  0.131 &  \textbf{54.8\%} & 0.031 \\
\bottomrule
\end{tabular}\label{tab:mixture}
\end{table}

\begin{table}[h]
\centering
\setlength{\tabcolsep}{1mm}
\caption{\neww{Ablation study of the stickman number on KIT dataset. IFG was not applied to save time.}}
\begin{tabular}{c c c c c}
\toprule
Stickman Number & FID $\downarrow$ & R-prec (top3) $\uparrow$ & StiSim $\uparrow$ & Diversity $\to$ \\
\midrule
0 & 0.171$^{\pm .015}$ & 0.799$^{\pm .012}$ & N/A & 10.76$^{\pm .147}$ \\
1 & 0.187$^{\pm .011}$ & 0.795$^{\pm .011}$ & 42.65\% & 10.77$^{\pm .144}$ \\
3 & 0.166$^{\pm .006}$ & 0.806$^{\pm .008}$ & 51.99\% & 10.79$^{\pm .148}$ \\
5 & 0.170$^{\pm .008}$ & 0.801$^{\pm .009}$ & 52.85\% & 10.79$^{\pm .149}$ \\
7 & \textbf{0.163$^{\pm .011}$} & 0.804$^{\pm .010}$ & \textbf{52.88\%} & 10.81$^{\pm .145}$ \\
9 & 0.168$^{\pm .007}$ & \textbf{0.805$^{\pm .011}$} & 52.67\% & 10.81$^{\pm .153}$ \\
\bottomrule
\end{tabular}
\label{tab:stickman}
\end{table}

\begin{table}[t]
\centering
\caption{\small{The time consumption of $repeat$ on the HumanML3D dataset with a batch size of 16.}}
\small
\renewcommand\arraystretch{1.2}
\setlength{\tabcolsep}{4.3mm}
\begin{tabular}{cccc}
\toprule
 $repeat$  & FID$\downarrow$ & Traj.err.$\downarrow$ &  Time/Batch (s)    \\ \toprule
10  &  0.137 & 0.062 & \textbf{7} \\ 
50 &  0.135 & 0.032 &  24 \\
100 &  \textbf{0.132} &  \textbf{0.026} & 44 \\
\bottomrule
\end{tabular}\label{tab:abl_repeat}
\end{table}

\definecolor{mygray}{gray}{.9}
\begin{table}[t]
\centering
\caption{\small{Comparison between stickman \& text-to-motion and text-to-motion task. ``TA" and ``TB" represent the time cost for overall and detailed descriptions, respectively, while ``TD" denotes the time required for hand-drawing. ``TI" represents the inference time of the utilized model. For Handmade animation, the trajectory is fixed and no textual input is required. all experiments are conducted on an A800 GPU with a batch size of 1. }}
\small
\renewcommand\arraystretch{1.2}
\setlength{\tabcolsep}{1.2mm}
\begin{tabular}{lccccccc}
\toprule
 Method & TA & TB & TD & TI & TotalTime$\downarrow$ & Score$\uparrow$    \\ \toprule
ReMoDiffuse~\cite{zhang2023remodiffuse} & 8.1s   &   24.5s  &  -   & 1.2s  & 33.8s & 7.3 \\
StickMotion~\cite{wang2025stickmotion} & 8.1s & -  & 7.7s  & 0.7s  & 16.4s & 8.5 \\
\rowcolor{mygray} DrawMotion & 8.1s & -  & 9.1s & 17.1s & 34.3s & 9.5 \\
Handmade & - & - & - & - & $\sim$3h & 7.4 \\
\bottomrule
\vspace{-10mm}
\end{tabular}\label{tab:user}
\end{table}

\noindent\textbf{Analysis on Condition Mixture.}\label{sec:abl_mixture}
As shown in Equation~\ref{eq:mixture}, the condition mixture controls how to combine the outputs based on four configurations of drawing and text conditions, which leads to a bias toward either drawing or text in the generated results. As discussed in Section~\ref{sec:diffusion}, $p(\hat{w}= w)$ determines the extent to which the coarse generations are biased toward the hand-drawing condition during the initial stage of the reverse process. The weights $(w_1,w_2,w_3,w_4)$ are then used to refine these coarse generations according to the four condition combinations in the final stage. The results across different configurations are relatively close, while the row with $p(\hat{w}=80\%)$ shows a stronger dependence on the drawing condition. To ensure a fair comparison and to reduce the burden on users to produce precise sketches, we adopt the configuration in the third row in our experiments.

\noindent\neww{\textbf{Analysis on the Number of Stickmen.}\label{sec:abl_stickmen}
Table~\ref{tab:stickman} presents an ablation study on the effect of the number of stickmen on the KIT dataset.
As the number of stickmen increases, the FID score exhibits slight fluctuations, with the best value observed at 7 stickmen.
R-Precision and StiSim generally improve with an increasing number of stickmen, although the gains become marginal beyond 3 stickmen.
The Diversity metric remains largely unchanged across different numbers of stickmen.
The limited impact on FID and Diversity can be attributed to the sparse temporal influence of stickmen, which affect only a small subset of frames.
Nevertheless, adding stickmen provides additional semantic cues, leading to modest improvements in text--motion alignment (R-Precision) and motion consistency (StiSim), particularly in sequences with minor motion variations, where multiple stickmen serve as mutual references to better capture user intent.
}

\noindent\textbf{Analysis on Intermediate Feature Guidance.}\label{sec:abl_ifg}
We have conducted parameter experiments on Intermediate Feature Guidance as shown in Table~\ref{tab:ma}, from which we can choose the best configuration. Moreover, Table~\ref{tab:abl_repeat} presents a time consumption analysis of $repeat$ times, demonstrating that improved performance can be achieved at the expense of increased computational time.

\noindent\textbf{Visualization of DrawMotion. }\label{sec:vis}
DrawMotion strives to meet users' demands as much as possible while ensuring the fidelity of the generated results. As shown in Figure~\ref{fig:vis}, for simple trajectory constraints such as (a), (d), and (e), the generation of DrawMotion can closely match the user's input. For more complex input trajectories of (b) and (c), the model attempts to ensure the vividness of the output without deviating too much from the input.

\noindent\textbf{User Study.} We recruited 20 independent volunteers to participate in the user study of DrawMotion, StickMotion, and the traditional text-to-motion work ReMoDiffuse~\cite{zhang2023remodiffuse}. These participants were instructed to imagine a specific human motion sequence lasting about 10 seconds and subsequently provide an overall textual description (A), detailed textual description (B), and the hand-drawing condition (D). The combination of (A, B) was inputted into ReMoDiffuse, while the combination of (A, D) was inputted into StickMotion and DrawMotion. Participants then rated these generated results on a scale of 0 to 10. Figure~\ref{fig:compare} shows the performance of the three methods and indicates that the output of DrawMotion is more aligned with the user's imagination. The final results regarding time consumption and user scores are shown in Table~\ref{tab:user}. 

\newx{Moreover, to further assess practical workflow efficiency, we additionally invited 5 professional animators to produce 3D stickman animations under the same trajectory constraints as our method. The aggregated results show that manual production requires about 3 hours per sample with a mean score of 7.4. We found that the handmade results adhere well to the target trajectory, but their motion naturalness is comparatively weaker. The animators further reported that AI-generated motions are richer and more natural than purely manual key-joint editing, and that DrawMotion's generation latency is acceptable in practice, whereas fully manual workflows remain considerably more time-consuming.}

DrawMotion saves users time in generating motions consistent with their imagination and achieves the highest level of satisfaction due to the precise control of the generations' trajectory.

\section{Limitation}~\label{sec:limit}
DrawMotion provides users with the highest level of creative freedom, allowing them to specify the trajectory and the character poses at designated positions along the trajectory. However, through our practice, we have found that when the trajectory or stickman figure input by the user conflicts with the text or violates fundamental principles of human motion, the generated motion sequence often deviates from the input, leading to reduced fidelity. In summary, while DrawMotion offers great flexibility, it also places the responsibility on users to ensure that their inputs are roughly physically reasonable and semantically consistent. \newm{Moreover, to indicate the degree of conflict, the final guidance loss, $||\hat{x}_{0}(\bar{F}, \dots) - c||_2^2$, from Algorithm~1 can be returned to the user to facilitate optimal configuration tuning.}

\section{Conclusion}~\label{sec:con}
This paper presents a novel hand-drawing condition and an efficient model DrawMotion for motion generation to address users' detailed requirements for generated motion through simple textual descriptions. To ensure consistency between the input conditions and generated motion sequences, we utilize both training-based and training-free guidance: The training-based guidance attempts to map the relationship between input and output through the model, where we introduce a stickman-based self-supervised encoding and an efficient Multi-Condition Module (MCM). The training-free guidance leverages the continuous intermediate feature space of MCM to receive gradients propagated from the classifier guidance, thereby further enhancing condition-generation alignment while preserving the fidelity of generation. In the experiments, we conduct both qualitative and quantitative analyses to validate the effectiveness of DrawMotion. Therefore, we firmly believe that our DrawMotion will be a professional and convenient motion generation method for art creators, and will promote the development of motion generation research and the relevant community.

\section*{Acknowledgment}
The paper is supported by National Natural Science Foundation of China No.62472046, No.62476224, and Young Elite Scientists Sponsorship Program of the Beijing High Innovation Plan No.20250866.

{
    \small
    \bibliographystyle{IEEEtran}
    \bibliography{main}
}

\begin{IEEEbiography}
[{\includegraphics[width=1in,height=1.25in,clip,keepaspectratio]{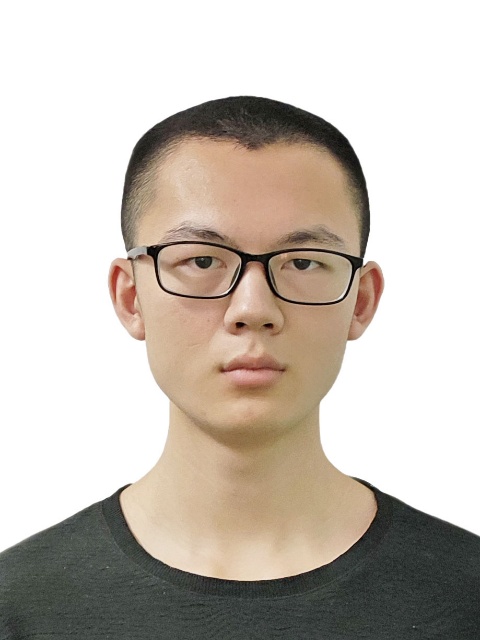}}]{Tao Wang} is currently pursuing a doctorate at Beijing University of Posts and Telecommunications (BUPT), Beijing, China. His major research areas include human pose estimation, refinement, generation, and editing, with related research results published in high-level conferences such as CVPR and ACMMM.
\end{IEEEbiography}

\begin{IEEEbiography}
[{\includegraphics[width=1in,height=1.25in,clip,keepaspectratio]{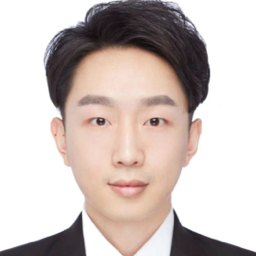}}]{Lei Jin} is currently an Associate Research Fellow with the Beijing University of Posts and Telecommunications (BUPT), Beijing, China. He graduated from Beijing University of Posts and Telecommunications. His major research areas include computer vision, data mining, and pattern recognition, with in-depth research in sub-fields such as human pose estimation, human action recognition, and human parsing, with related research results published in high-level conferences and journals such as CVPR, AAAI, NIPS, IJCAI, and ACMMM, and so on.
\end{IEEEbiography}

\begin{IEEEbiography} [{\includegraphics[width=1in,height=1.25in,clip,keepaspectratio]{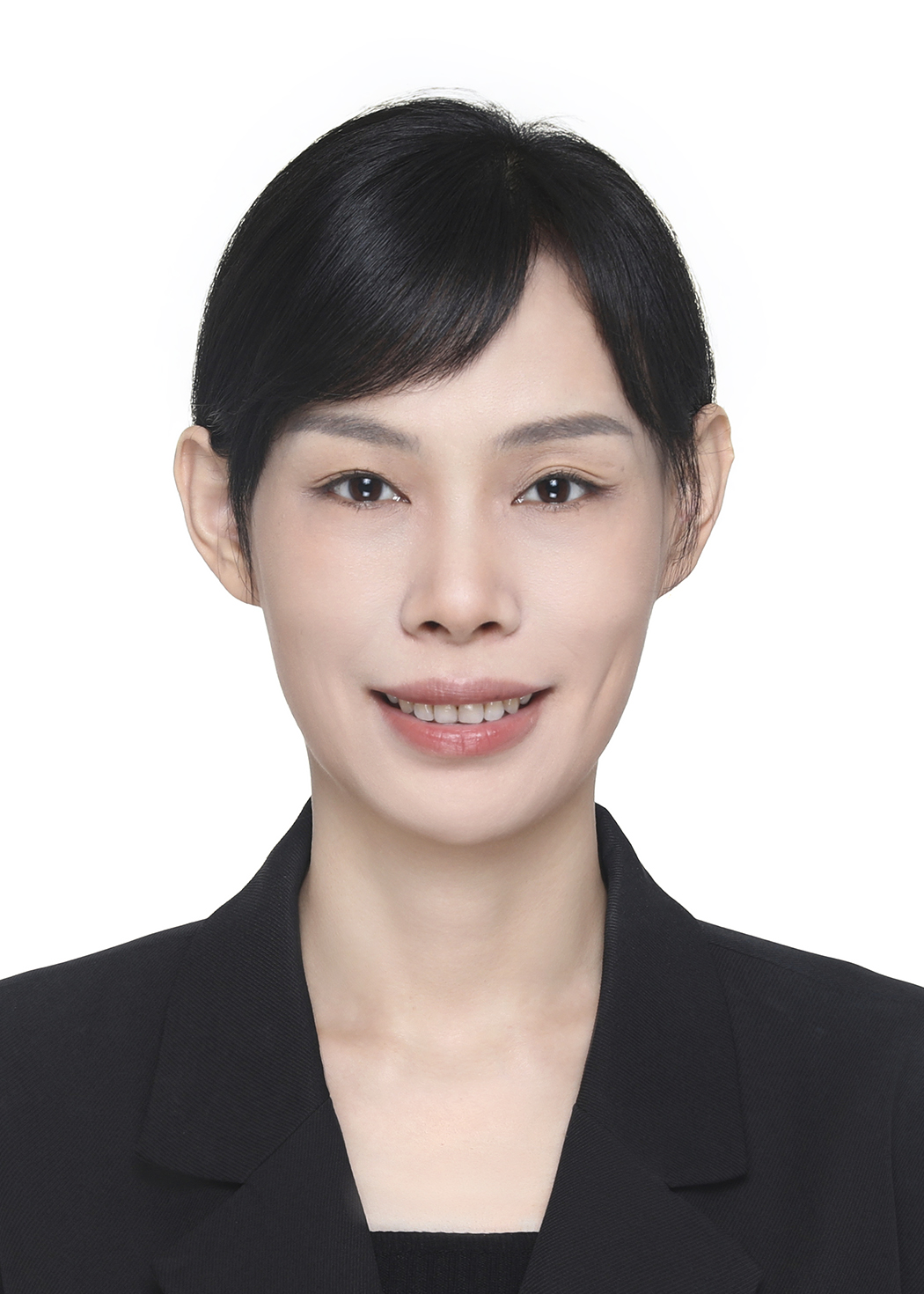}}]{Zhihua Wu} graduated from the University of Science and Technology of China and has been engaged in long-term research in artificial intelligence. Her work focuses on large language model training and acceleration, as well as human motion analysis and related applications.
\end{IEEEbiography}

\begin{IEEEbiography} [{\includegraphics[width=1in,height=1.25in,clip,keepaspectratio]{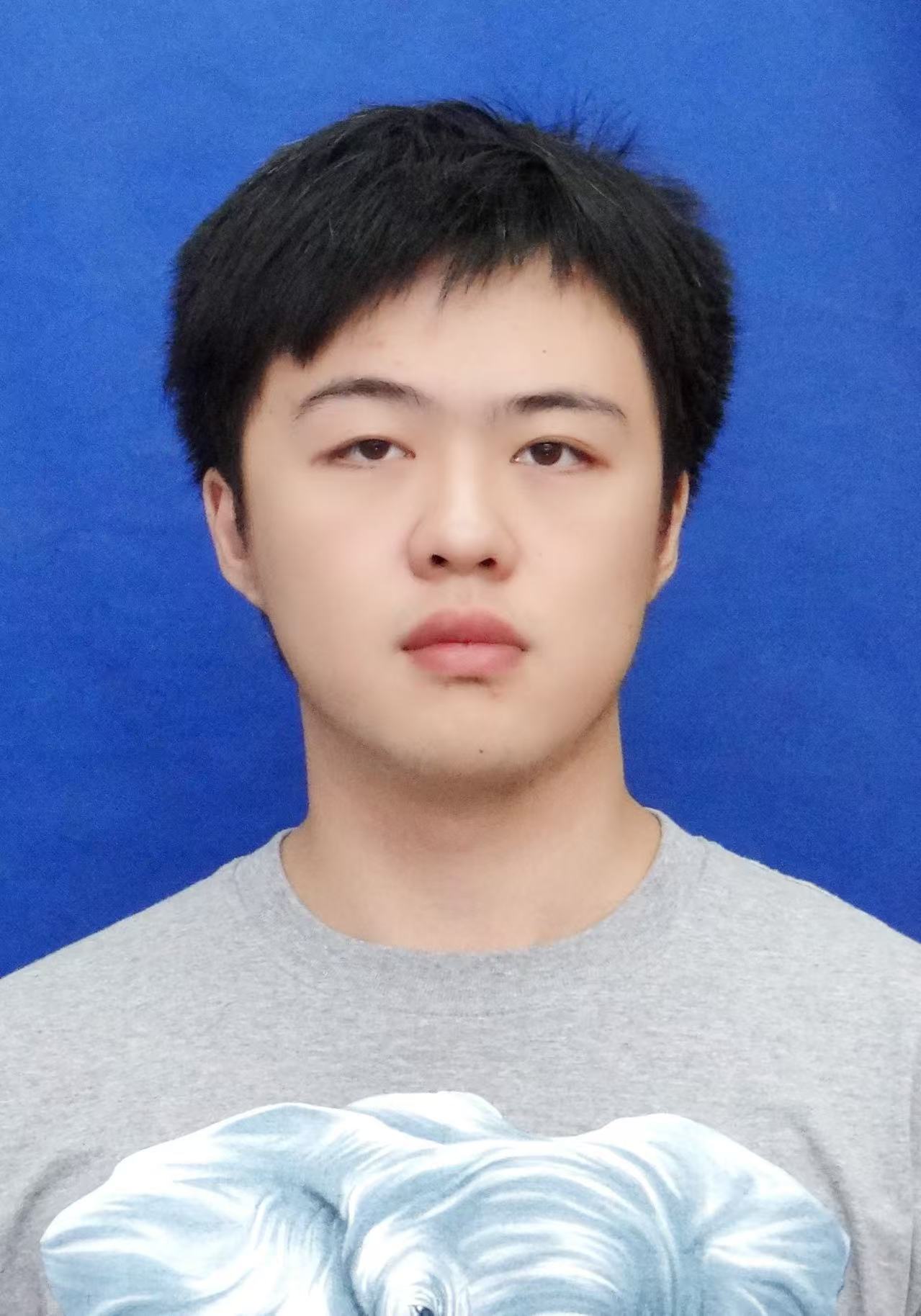}}]{Qiaozhi He} is a researcher focusing on large language models (LLMs) and natural language processing. His work covers LLM training, inference optimization, reward modeling, and multimodal learning. He has coauthored papers in leading venues such as AAAI and CVPR, including studies on cross-layer attention sharing, efficient inference, and human motion generation. His current interests include scalable model design, efficient deployment, and preference-aligned AI systems.
\end{IEEEbiography}

\begin{IEEEbiography} [{\includegraphics[width=1in,height=1.25in,clip,keepaspectratio]{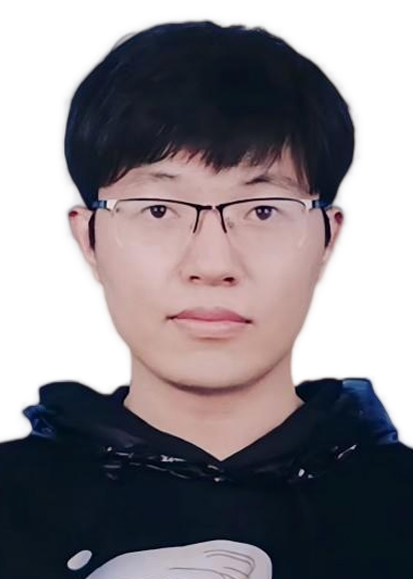}}]{Jiaming Chu} is currently pursuing the Ph.D. degree in electronic science and technology at Beijing University of Posts and Telecommunications. His research interests include deep learning and computer vision, with in-depth research in sub-fields, such as human action recognition, instance segmentation, human parsing, and diffusion model.
\end{IEEEbiography}

\begin{IEEEbiography} [{\includegraphics[width=1in,height=1.25in,clip,keepaspectratio]{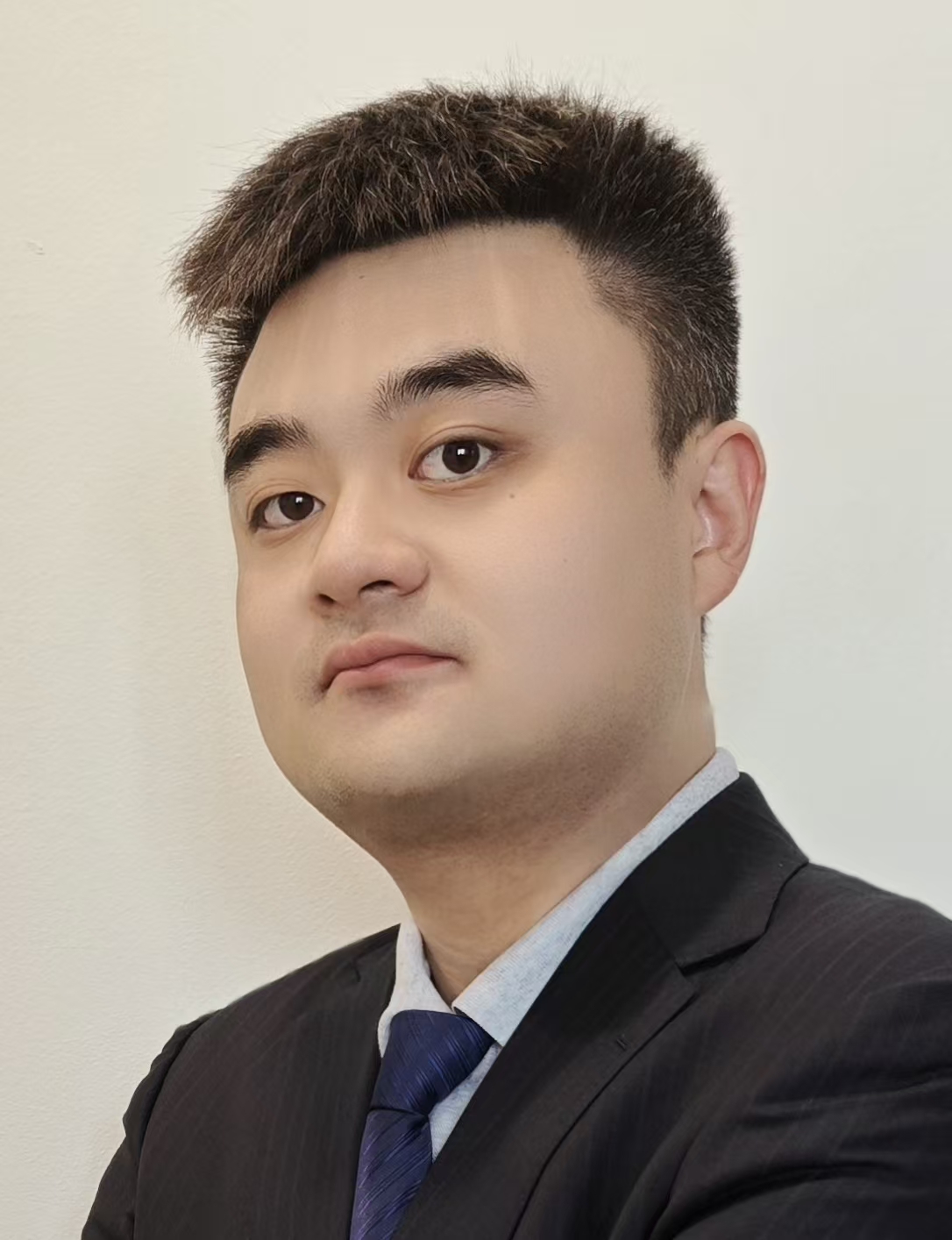}}]{Yu Cheng} received his Ph.D. degree in Electrical and Computer Engineering from the University of Singapore. His research interests include human pose estimation, facial recognition and object detection. He has published papers in top conferences such as CVPR, ECCV, etc.
\end{IEEEbiography}

\begin{IEEEbiography} [{\includegraphics[width=1in,height=1.25in,clip,keepaspectratio]{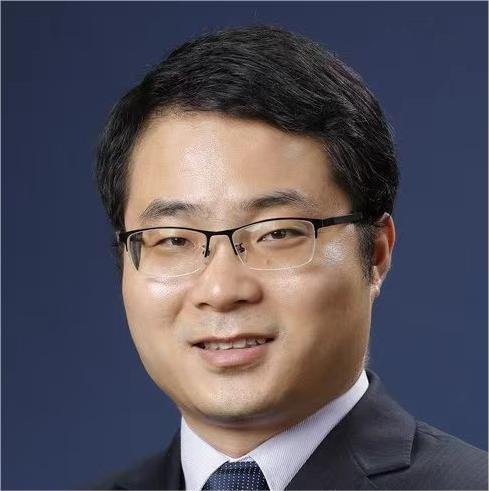}}]{Junliang Xing} is a Professor with Tsinghua University and the recipient of the National Science Fund for Excellent Young Scholar. 
He obtained his dual bachelor's degrees in Computer Science and Mathematics at Xi'an Jiaotong University in 2007 and his doctorate in Computer Science in 2012. Then he worked in the National Laboratory of Pattern Recognition, Institute of Automation, Chinese Academy of Sciences as an assistant researcher, associated researcher, and researcher in 2012, 2015, and 2018, respectively. 
His research interests are human-computer interactive learning, computer gaming, and computer vision. He has published more than 100 peer-reviewed papers in international conferences and journals and got more than 13,000 citations from Google Scholar. 
\end{IEEEbiography}

\begin{IEEEbiography} [{\includegraphics[width=1in,height=1.25in,clip,keepaspectratio]{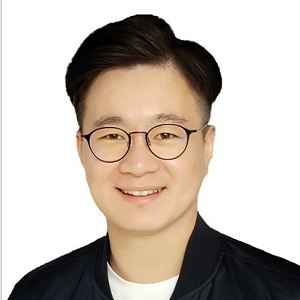}}]{Jian Zhao} is the leader of Evolutionary Vision+x Oriented Learning (EVOL) Lab and Young Scientist at Institute of AI (TeleAI), China Telecom, and Researcher and Ph.D. Supervisor at Northwestern Polytechnical University (NWPU). He received his Ph.D. degree from National University of Singapore (NUS) in 2019 under the supervision of Assist. Prof. Jiashi Feng and Assoc. Prof. Shuicheng Yan. He is the SAC of VALSE, the committee member of CSIG-BVD, and the member of the board of directors of BSIG. He has over 40 influential papers on human-centric image understanding, and accolades including the Lee Hwee Kuan Gold and ACM MM Best Student Paper awards.  Additionally, he has organized key workshops and challenges at CVPR, ECCV, and other venues.
\end{IEEEbiography}

\begin{IEEEbiography} [{\includegraphics[width=1in,height=1.25in,clip,keepaspectratio]{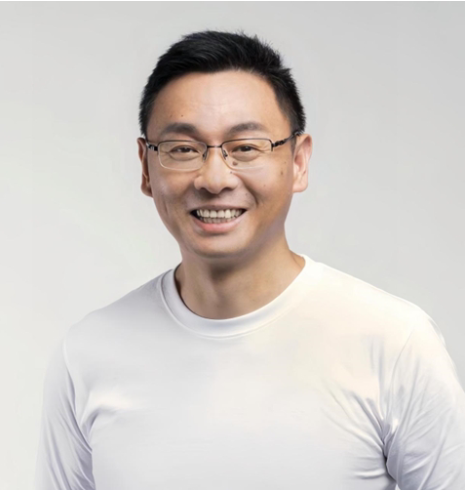}}]{Shuicheng Yan} (Fellow, IEEE) is Managing Director of Kunlun 2050 Research and Chief Scientist of Kunlun Tech \& Skywork AI, and formerly Group Chief Scientist of Sea. He is a Fellow of the Singapore Academy of Engineering, AAAI, ACM, IEEE, and IAPR. His research focuses on computer vision, machine learning, and multimedia analysis. Prof. Yan has published over 800 papers with an H-index above 140 and has been named a World’s Highly Cited Researcher nine times. His team has achieved top awards at major competitions such as Pascal VOC and ImageNet (ILSVRC), and has won multiple best paper and best student paper awards, including several at ACM Multimedia.
\end{IEEEbiography}

\begin{IEEEbiography} [{\includegraphics[width=1in,height=1.25in,clip,keepaspectratio]{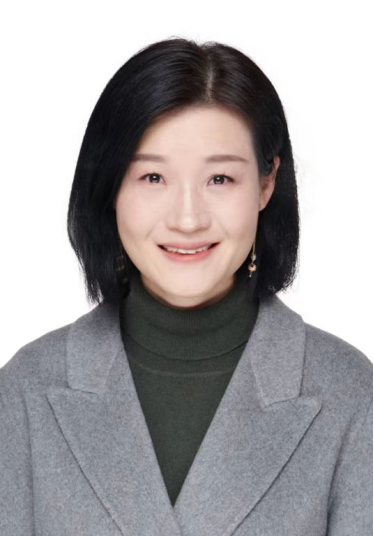}}]{Li Wang} (Senior Member, IEEE) received her Ph.D. from BUPT in 2009 and is now a Full Professor at the School of Electronic Engineering, BUPT, where she leads the High Performance Computing and Networking Lab and serves as Associate Dean of the School of Software Engineering. She has held visiting positions at Georgia Tech and Chalmers University. Her research interests include wireless communications, distributed networking, vehicular communications, social networks, and edge AI. She has authored nearly 50 journal papers, two books, and received multiple best paper awards. She also serves on editorial boards of several IEEE and international journals.
\end{IEEEbiography}

\end{document}